\documentclass[journal,twoside,web]{ieeecolor}
\usepackage{generic}
\usepackage{cite}
\usepackage{amsmath,amssymb,amsfonts}
\usepackage{graphicx}
\usepackage{hyperref}
\hypersetup{hidelinks=true}
\usepackage{textcomp}

\usepackage[ruled]{algorithm2e}
\usepackage{bm}
\usepackage[caption=false,subrefformat=parens]{subfig}
\usepackage{booktabs}
\usepackage{threeparttable}
\usepackage{multirow}
\usepackage{multicol}
\usepackage{courier}

\def\R{\mathbb{R}}
\def\calX{\mathcal{X}}
\def\rmX{\mathrm{X}}

\def\pmb#1{\mathbf{#1}}

\newtheorem{definition}{Definition}[section]
\newtheorem{theorem}{Theorem}[section]
\newtheorem{proposition}{Proposition}[section]
\newtheorem{lemma}{Lemma}[section]
\newtheorem{example}{Example}[section]

\def\BibTeX{{\rm B\kern-.05em{\sc i\kern-.025em b}\kern-.08em
    T\kern-.1667em\lower.7ex\hbox{E}\kern-.125emX}}
\begin{document}
\title{Feasible Policy Iteration for Safe Reinforcement Learning}
\author{Yujie Yang, Zhilong Zheng, Shengbo Eben Li, Wei Xu, Jingjing Liu, Xianyuan Zhan, Ya-Qin Zhang
% \thanks{This paragraph of the first footnote will contain the date on 
% which you submitted your paper for review. It will also contain support 
% information, including sponsor and financial support acknowledgment. For 
% example, ``This work was supported in part by the U.S. Department of 
% Commerce under Grant 123456.'' }
\thanks{Yujie Yang and Zhilong Zheng contributed equally to this work. All correspondence should be sent to Shengbo Eben Li.}
\thanks{Yujie Yang, Zhilong Zheng, and Shengbo Eben Li are with the School of Vehicle and Mobility, Tsinghua University, Beijing, 10084, China (e-mail: yangyj21@mails.tsinghua.edu.cn, zheng-zl22@mails.tsinghua.edu.cn, lishbo@tsinghua.edu.cn).}
\thanks{Wei Xu is with the College of Artificial Intelligence, Tsinghua University, Beijing, 10084, China (e-mail: weixu@tsinghua.edu.cn).}
\thanks{Jingjing Liu, Xianyuan Zhan, and Ya-Qin Zhang are with the Institute for AI Industry Research, Tsinghua University, Beijing, 10084, China (e-mail: jjliu@air.tsinghua.edu.cn, zhanxianyuan@air.tsinghua.edu.cn, zhangyaqin@air.tsinghua.edu.cn).}
}

\maketitle

\begin{abstract}
Safety is the priority concern when applying reinforcement learning (RL) algorithms to real-world control problems.
While policy iteration provides a fundamental algorithm for standard RL, an analogous theoretical algorithm for safe RL remains absent.
In this paper, we propose feasible policy iteration (FPI), the first foundational dynamic programming algorithm for safe RL.
FPI alternates between policy evaluation, region identification and policy improvement.
This follows actor-critic-scenery (ACS) framework where scenery refers to a feasibility function that represents a feasible region.
A region-wise update rule is developed for the policy improvement step, which maximizes state-value function inside the feasible region and minimizes feasibility function outside it.
With this update rule, FPI guarantees monotonic expansion of feasible region, monotonic improvement of state-value function, and geometric convergence to the optimal safe policy.
Experimental results demonstrate that FPI achieves strictly zero constraint violation on low-dimensional tasks and outperforms existing methods in constraint adherence and reward performance on high-dimensional tasks.
\end{abstract}


\section{Introduction}
\IEEEPARstart{R}{inforcement} learning has achieved promising performance on many challenging tasks such as video games~\cite{vinyals2019grandmaster}, board games~\cite{schrittwieser2020mastering}, robotics~\cite{andrychowicz2020learning}, and autonomous driving~\cite{guan2022integrated}. RL solves an optimal control problem (OCP) by finding a policy that maximizes the expected cumulative rewards. However, many real-world control tasks require not only maximizing rewards but also satisfying safety constraints. Such problems can be formulated as constrained OCPs~\cite{li2023reinforcement}. In this paper, we consider stepwise deterministic constraints, which require strict constraint satisfaction at every step.

Policy iteration (PI) is a fundamental dynamic programming algorithm for unconstrained RL~\cite{sutton2018reinforcement}.
It iteratively performs two steps: policy evaluation (PEV), which computes the state-value function of the current policy, and policy improvement (PIM), which updates the policy by maximizing the state-value function.
% PI ensures monotonic improvement of the state-value function and guarantees geometric convergence to the optimal policy.
PI lays the theoretical foundation for many modern RL algorithms~\cite{lillicrap2015continuous, fujimoto2018addressing, haarnoja2018soft}.
% , such as deep deterministic policy gradient (DDPG)~\cite{lillicrap2015continuous}, twin delayed DDPG (TD3)~\cite{fujimoto2018addressing}, soft actor-critic (SAC)~\cite{haarnoja2018soft}, etc.
Despite its success, PI is limited to unconstrained RL and cannot handle safety constraints.
In the context of safe RL, a foundational algorithm analogous to PI remains undeveloped.

Existing safe RL methods essentially combine constrained optimization techniques with RL algorithms.
A prominent class of methods is called \textit{iterative unconstrained RL}, which reformulates the safe RL problem as a sequence of unconstrained optimization problems and solves each one with standard RL algorithms.
The most widely used reformulation method is the Lagrange multiplier method~\cite{boyd2004convex}, where constraints are added to the objective function with Lagrange multipliers to form an unconstrained optimization problem.
For specific types of constraints, such as the cost value function in the constrained Markov decision process (CMDP), the Lagrange function can be rearranged into a discounted summation.
With this rearrangement, the unconstrained problem turns out to be an RL problem itself and can be solved using any RL algorithm.
The problem with iterative unconstrained RL methods is that they suffer from slow convergence and unstable training.
The slow convergence arises from the need to solve an RL problem in every iteration of dual ascent, which, in theory, makes these algorithms converge a magnitude slower than standard RL algorithms.
% In practice, the inner RL problems are only approximately solved with a limited number of updates, further complicating the convergence of dual ascent.
The instability in training stems from the inherent characteristic of the Lagrange method, which only guarantees convergence at the end but provides no assurance for intermediate objective functions or constraints.
In safe RL, this lack of intermediate guarantee is manifested as oscillations of reward performance and constraint violations during training.

\begin{figure*}
    \centering
    \subfloat[PI]{
        \includegraphics[width=0.3\linewidth, trim=20 15 25 15, clip]{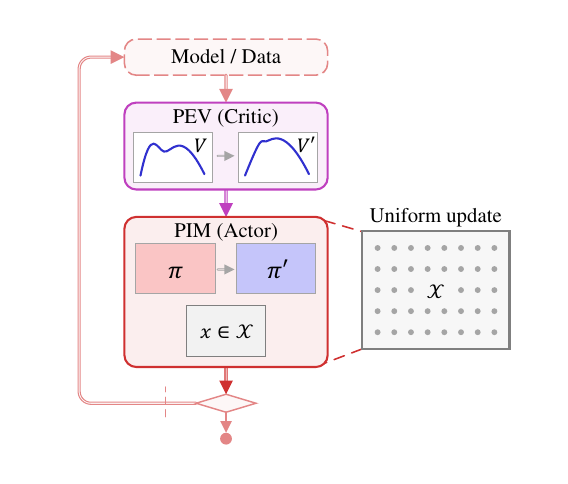}
    }
    \subfloat[FPI]{
        \includegraphics[width=0.36\linewidth, trim=25 15 20 15, clip]{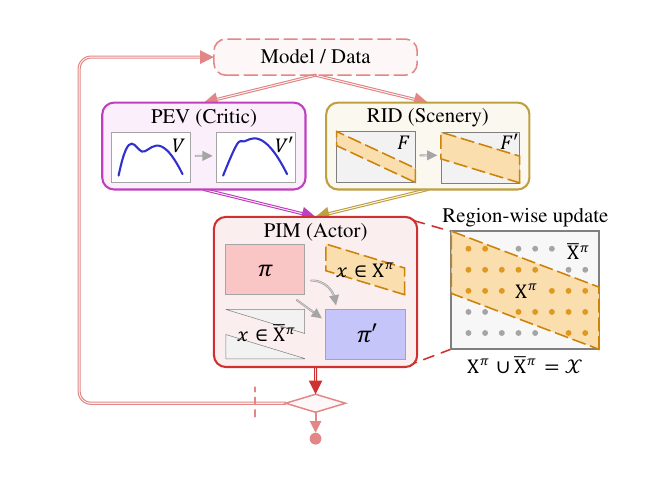}
    }
    \hspace{2pt}
    \subfloat[FPI convergence mechanism]{
        \includegraphics[width=0.27\linewidth, trim=20 15 20 15, clip]{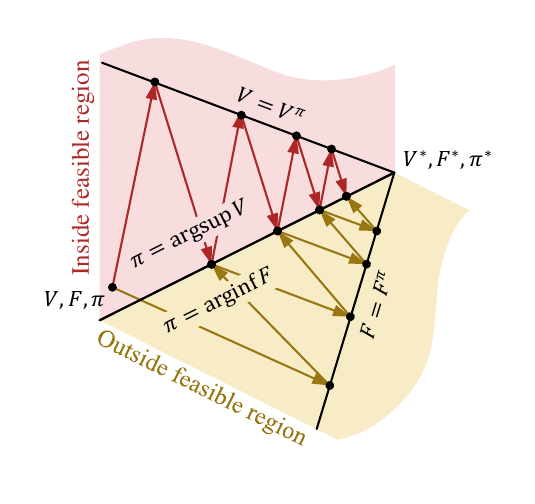}
    }
    \caption{Iteration framework and convergence mechanism of FPI. (a) PI adopts the AC framework. (b) The ACS framework for FPI features an additional RID step and a region-wise PIM. (c) The iteration of FPI converges to the optimal solution of safe RL.}
    \label{fig: PI vs FPI}
\end{figure*}

\begin{figure}[htbp]
    \centering
    \includegraphics[width=\linewidth, trim=18 20 20 18, clip]{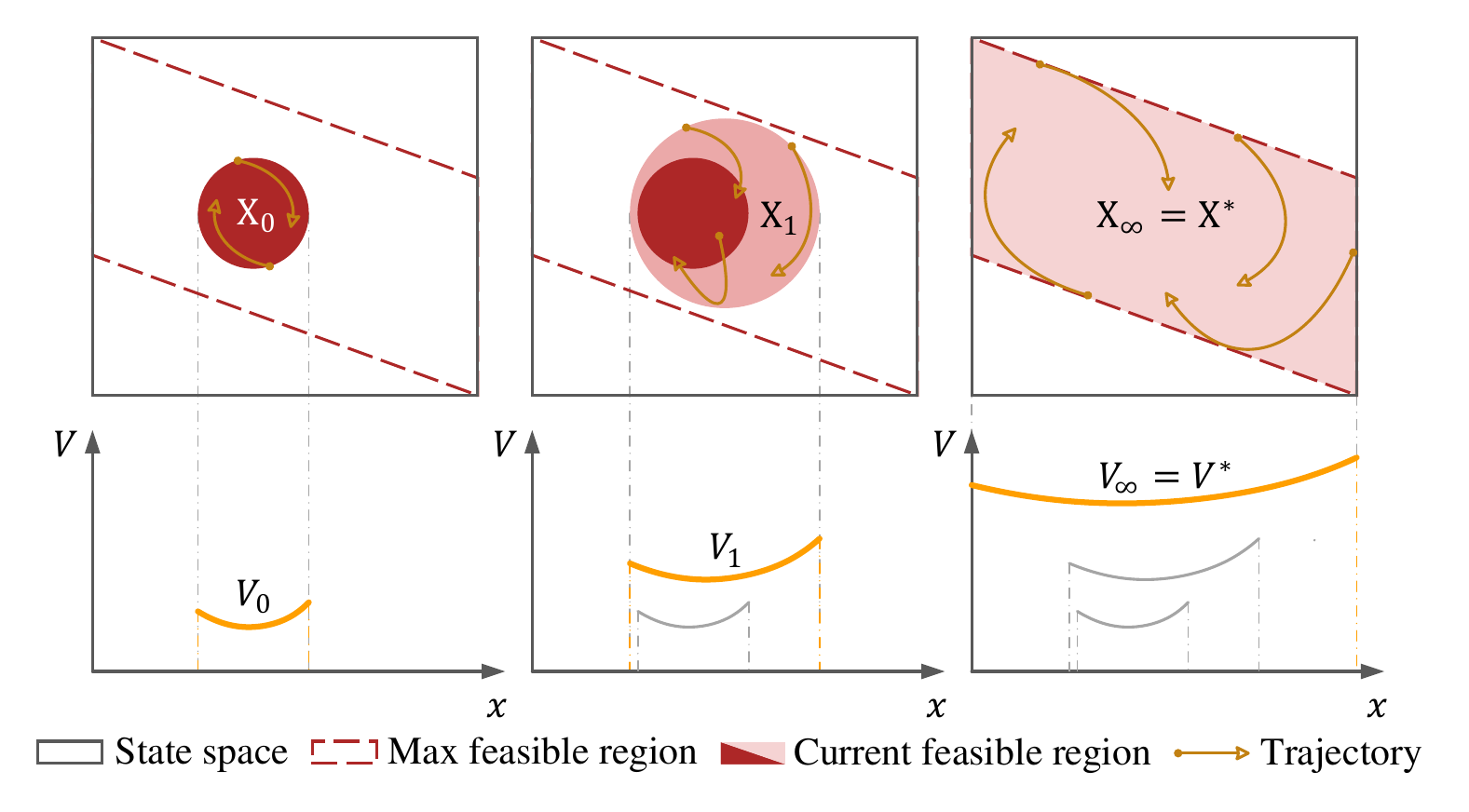}
    \caption{Monotonicity and convergence of FPI. The feasible region monotonically expands and converges to the maximum feasible region. The state-value function monotonically increases inside the feasible region and converges to the optimal state-value function.}
    \label{fig: monotonicity}
\end{figure}

Another class of methods is called \textit{constrained policy optimization}, which follows the policy optimization framework of standard RL and incorporates safety constraints into each iteration.
Policy optimization methods treat RL as an optimization problem and solve it using established optimization algorithms.
For example, vanilla policy gradient uses a gradient descent method, and trust region policy optimization (TRPO)~\cite{schulman2015trust} uses a trust region method.
These optimization algorithms typically involve an iteration process.
In safe RL, the constraint can be incorporated into these iterations.
The most representative example is constrained policy optimization (CPO)~\cite{achiam2017constrained}, which adapts TRPO's trust region update framework by adding a safety constraint in each iteration.
The constraint function is linearized within the trust region to obtain an analytical solution.
The problem with these methods is that they suffer from the infeasibility issue, i.e., they may fail to find a constraint-satisfying solution in some iterations.
This is because the constraint, which remains fixed throughout training, is too stringent, especially at an early stage.
At the beginning of training, the policy is usually randomly initialized and requires numerous updates before it can satisfy the safety constraint.
When infeasibility occurs, these methods resort to alternative update rules, e.g., solely minimizing the constraint function~\cite{achiam2017constrained}.
This prevents the algorithm from optimizing rewards until the constraints are fully met, resulting in overly conservative behavior and low training efficiency.

The aforementioned challenges of existing safe RL algorithms stem from the lack of a theoretically effective and efficient iteration approach for safe RL that is analogous to PI for standard RL.
To this end, we propose the first foundational dynamic programming algorithm for safe RL called feasible policy iteration (FPI).
Different from PI which adopts an actor-critic (AC) framework, FPI follows an actor-critic-scenery (ACS) framework, which additionally includes a region identification (RID) step for updating the ``scenery" and performs a region-wise update in PIM, as shown in Fig. \ref{fig: PI vs FPI}.
The ``scenery" in ACS refers to a feasibility function that represents the long-term constraint-satisfying set of a policy.
% FPI also differs from PI in that it features a region-wise update over the state space, while the latter performs a uniform update.
% In FPI, PEV and RID compute the state-value function and feasibility function of the current policy respectively, and 
Region-wise PIM updates the policy by minimizing the feasibility function outside the feasible region and maximizing the state-value function inside the feasible region.
By recursively applying the self-consistency conditions, we prove that FPI guarantees monotonic expansion of the feasible region and monotonic increase of the state-value function within the feasible region.
Furthermore, by deriving the contraction properties of two Bellman operators, we prove that both the feasibility function and state-value function geometrically converge to their optimum, as illustrated in Fig. \ref{fig: monotonicity}.
% To deal with continuous state and action spaces, we develop an algorithm based on FPI that approximates the feasibility function, state-value function, and policy using neural networks and optimizes them via gradient descent.
Experiment results show that FPI can learn strictly safe and near-optimal policies on low-dimensional tasks, and it can also achieve fewer violations and better performance on high-dimensional tasks.

\section{Related Work}
% Existing safe RL methods can be divided into two classes: iterative unconstrained RL and constrained policy optimization.
% Iterative unconstrained RL methods reformulate safe RL as a sequence of unconstrained optimization problems and solve them using standard RL algorithms.
% Constrained policy optimization methods follow the policy optimization framework and add safety constraints to every update step.

\subsection{Iterative unconstrained RL methods}
Most iterative unconstrained RL methods use the Lagrange method and solve the dual problem using dual ascent, where the minimization step involves an unconstrained RL problem.
The theoretical basis for these methods is established by Paternain et al.~\cite{paternain2019constrained}, who prove that the safe RL problem has zero duality gap.
Under this framework, different types of constraints can be used.
For example, Chow et al.~\cite{chow2017risk} constrain the conditional value-at-risk of the cumulative cost, forming a probabilistic constraint.
Tessler et al.~\cite{tessler2019reward} add the cost signal to the reward and treat the combined discounted sum as a new value function.
Yu et al.~\cite{yu2023safe} learn a Hamilton-Jacobi (HJ) reachability function and constrain the policy in its reachable set.
Several works have also employed HJ reachability functions as safety constraints~\cite{fisac2019bridging, hsu2021safety, yu2022reachability, ganai2024iterative}.
Besides HJ reachability function, other safety certificates frequently used in safe RL include control barrier function (CBF)~\cite{ma2021model, wang2021safe, xu2022adaptive, yang2023model, mu2024safe, yang2024synthesizing}, safety index~\cite{liu2014control, ma2022joint}, and Lyapunov function~\cite{richards2018lyapunov, chow2018lyapunov, chang2019neural}.
For example, Yang et al.~\cite{yang2023model} learn a neural CBF and construct a multi-step barrier constraint for policy update.
Ma et al.~\cite{ma2022joint} learn a safety index with a given functional form and adjustable parameters by minimizing energy increase.
Chow et al.~\cite{chow2018lyapunov} construct a Lyapunov function to guarantee the global safety of the policy via a set of local linear constraints.
The limitation of iterative unconstrained RL methods is that their convergence is slow because they introduce an additional iterative loop on top of standard RL.
Moreover, their learning process exhibit significant oscillation~\cite{stooke2020responsive} because of the lack of intermediate guarantees of the Lagrange method.
While empirical techniques such as PID-Lagrangian~\cite{stooke2020responsive, peng2022model} are proposed to reduce oscillations, they still cannot provide theoretical guarantees for intermediate policies.

\subsection{Constrained policy optimization methods}
The most representative algorithm of this class is CPO~\cite{achiam2017constrained}, which builds on ideas of trust region update and Taylor expansion approximation from TRPO~\cite{schulman2015trust}.
Within this framework, CPO further approximates the constraint of the cost value function with an affine function and analytically solves the resulting constrained optimization problem.
The authors prove that CPO ensures a lower bound on policy performance and an upper bound on constraint violation in every iteration.
To avoid the computationally expensive line search in CPO, Yang et al.~\cite{yang2020projection} propose projection-based policy optimization (PCPO), which first performs a reward improvement update and then projects the policy back onto the constrained set.
Zhang et al.~\cite{zhang2020first} propose first-order constrained optimization in policy space (FOCOPS), which first solves for the optimal update policy in the non-parameterized policy space and then projects it back into the parametric policy space.
Following the policy projection route, Yang et al.~\cite{yang2022constrained} develop constrained update projection (CUP), which generalizes the surrogate functions estimator and only requires first-order optimizers without relying on any strong approximation on the convexity of the objectives.
Inspired by the interior-point method, Liu et al.~\cite{liu2020ipo} incorporate a logarithmic barrier function of the constraint to the objective function and optimize it using a first-order policy optimization method.
Additionally, several algorithms integrate the Lagrange multiplier method with policy optimization, such as proximal policy optimization with Lagrangian (PPO-Lagrangian)~\cite{ray2019benchmarking}, TRPO-Lagrangian~\cite{ray2019benchmarking}, and augmented PPO~\cite{dai2023augmented}, also falling under this category.
A limitation of these methods is their strict requirement for the policy to satisfy the original safety constraint in all iterations.
% This constraint is often too stringent, especially during early training stages.
Enforcing such constraints can lead to infeasibility issues, as mentioned by Achiam et al.~\cite{achiam2017constrained}.
It also causes overly conservative behavior in an early training stage, as the policy must focus solely on minimizing constraint violations without improving performance.

\section{Problem Formulation}
\subsection{Markov decision process with state constraints}
In this paper, we consider a deterministic Markov decision process (MDP) described by a 5-tuple $(\mathcal{X},\mathcal{U},f,r,\gamma)$, where $\mathcal{X}\subseteq\mathbb{R}^n$ is the finite state space, $\mathcal{U}\subseteq\mathbb{R}^m$ is the finite action space, $f: \mathcal{X}\times\mathcal{U}\to\mathcal{X}$ is the dynamics model, $r: \mathcal{X}\times\mathcal{U}\to\mathbb{R}$ is the reward function, $0<\gamma<1$ is the discount factor.
The policy in the MDP considered in this paper is a deterministic function that maps a state to an action: $\pi:\mathcal{X}\to\mathcal{U}$.
The state-value function under a policy $\pi$, $V^\pi: \calX\to\R$, is defined as the discounted sum of rewards under $\pi$:
\begin{equation}
    V^\pi(x_0)=\sum_{t=0}^\infty\gamma^t r(x_t,\pi(x_t)),\forall x_0\in\mathcal{X},
\end{equation}
where $x_{t+1}=f(x_t,\pi(x_t))$.
This function is the objective we aim to maximize in RL. Note that since we are considering a finite MDP, a state-value function $V^\pi$ can also be viewed as a vector in $\R^{|\calX|}$. This perspective can be convenient in the following analysis, so we will interchangeably refer to $V^\pi$ as a function and a vector in the rest of the paper.

Safety is specified through state constraints in the form of an inequality $h(x)\leq0$, where $h: \mathcal{X}\to\mathbb{R}$ is the constraint function.
The constraint function gives an instantaneous safety requirement that should be satisfied at every time step, i.e., we require that
\begin{equation}
\label{eq: infinite-horizon constraints}
    h(x_t)\le0,\forall t\ge0.
\end{equation}
However, satisfying these infinite-horizon constraints is not always possible.
Starting from some states, we may not be able to find a long-term safe policy, and the state constraint will inevitably be violated at some time step in the future.
To formally describe and study the long-term safety under state constraints, we introduce the concept of feasibility.

\subsection{Feasibility in safe reinforcement learning}
Feasibility in safe RL includes state feasibility and policy feasibility.
State feasibility refers to the satisfiability of the infinite-horizon state constraints, while policy feasibility describes the long-term safety of a specific policy.

\begin{definition}[State feasibility]
\label{def: state feasibility}~
\begin{enumerate}
    \item A state $x_0\in\mathcal{X}$ is feasible if there exists a policy $\pi$, such that for all $t\ge0$, we have $h(x_t)\le0$, where $x_{t+1}=f(x_t,\pi(x_t))$.
    \item A set $\mathrm{X}\subseteq\mathcal{X}$ is a feasible region if all states in this set are feasible.
    \item The maximum feasible region $\mathrm{X}^*\subseteq\mathcal{X}$ is the set of all states that are feasible.
\end{enumerate}
\end{definition}

By definition of state feasibility, the infinite-horizon constraints \eqref{eq: infinite-horizon constraints} can be satisfied if and only if the initial state $x_0$ is in the maximum feasible region.
Given a policy in the MDP, some initial states in the maximum feasible region may satisfy the infinite-horizon constraints while others may not.
Since safety is our priority concern, we require that the policy we find should render all states in the maximum feasible region safe in the long term.
To formally describe this requirement, we introduce the concept of policy feasibility.

\begin{definition}[Policy feasibility]
\label{def: policy feasibility}~
\begin{enumerate}
    \item A policy $\pi$ is feasible in a state $x_0\in\mathcal{X}$ if for any $t\ge0$, we have $h(x_t)\le0$, where $x_{t+1}=f(x_t,\pi(x_t))$.
    \item The feasible region of $\pi$, $\mathrm{X}^\pi\subseteq\mathcal{X}$, is the set of all states in which $\pi$ is feasible.
\end{enumerate}
\end{definition}

The feasible region of policy is by definition a subset of the maximum feasible region, i.e., for any $\pi$, we have $\mathrm{X}^\pi\subseteq\mathrm{X}^*$.
The requirement that a policy renders all states in the maximum feasible region safe in the long term is equivalent to the fact that the feasible region of this policy equals the maximum feasible region.
Another way to understand this requirement is through the working area of a policy.
The feasible region of a policy is its safe working area, i.e., the set where it can be safely applied.
Due to the safety priority requirement, we want this area to be as large as possible.
Furthermore, a larger feasible region may also lead to better control performance because it allows the policy to explore a larger set in the state space, increasing the possibility of obtaining higher rewards.

\subsection{Objective of safe reinforcement learning}
Through the above analysis, we know that the safety requirement in safe RL is to find a policy that has the maximum feasible region.
Together with the performance objective of maximizing the state-value function, we arrive at the objective of safe RL:
\begin{equation}
\label{eq: safe RL objective}
    \max_{\pi\in\Pi^*}V^\pi(x),\forall x\in\mathrm{X}^*,
\end{equation}
where $\Pi^*$ is the set of policies with the maximum feasible region, i.e., $\Pi^*=\{\pi|\mathrm{X}^\pi=\mathrm{X}^*\}$.

\section{Feasibility Function and Optimality Condition}

\subsection{Notation declaration}
For simplicity of analysis, we declare the following notation usages.
\begin{itemize}
    % \item When using the order relation operators $\le$, $\ge$, and $=$ between functions, we refer to the pointwise comparison.
    \item Given a non-empty subset $\rmX\subseteq\calX$, we define an equivalence relation ${=_\rmX}$ and a partial order ${\le_\rmX}$ on $\R^{|\calX|}$:
    $$
    \begin{aligned}
        V_1 =_\rmX V_2\iff V_1(x)= V_2(x),\forall x\in\mathrm{X}, \\
        V_1 \le_\rmX V_2\iff V_1(x)\le V_2(x),\forall x\in\mathrm{X},
    \end{aligned}
    $$
    where $=$ and $\le$ are correspondingly the normal equivalence relation and partial order on $\R$. It holds directly from the properties of $=$ and $\le$ that the definitions of ${=_\rmX}$ and ${\le_\rmX}$ are valid. We will omit the subscript when $\rmX=\calX$ since ${=_\rmX}$ and ${\le_\rmX}$ degenerate correspondingly to the normal pointwise equivalence and partial order on $\R^{|\calX|}$ under such cases.
    \item Given a non-empty subset $\rmX\subseteq\calX$, we construct a normed space $\left(\R^{|\calX|}, =_\rmX, \le_\rmX, \Vert\cdot\Vert_{\mathrm{X},\infty}\right)$ by equipping $\left(\R^{|\calX|}, =_\rmX, \le_\rmX \right)$ with a norm $\Vert\cdot\Vert_{\mathrm{X},\infty}$:
    $$\Vert V\Vert_{\mathrm{X},\infty}=\sup_{x\in\mathrm{X}}|V(x)|, \forall V\in\R^{|\calX|}.$$
    This is equivalent to the normal infinity norm $\Vert\cdot\Vert_{\infty}$ on $\R^{|\calX|}$ when $\rmX=\calX$.
    \item We use $(\cdot)_+$ to denote clipping a function to non-negative values, i.e., $(\cdot)_+=\max\{\cdot,0\}$.
    \item We use $\bar{\cdot}$ to denote the complement of a set, e.g., $\bar{\mathrm{X}}=\mathcal{X}\setminus\mathrm{X}$.
\end{itemize}

\subsection{Feasibility function}
Solving problem \eqref{eq: safe RL objective} is challenging because the feasible region of a policy is difficult to represent, and it is even more difficult to decide whether a feasible region equals the maximum one.
This difficulty stems from the definition of feasibility, which involves an infinite number of constraints and is thus almost impossible to check directly.
To solve this problem, we introduce a tool called the feasibility function that directly represents a feasible region with its zero-sublevel set.
This function also satisfies several desirable properties so that it can be easily computed.

\begin{definition}[Feasibility function]
\label{def: feasibility function}
$F^\pi: \mathcal{X}\to\mathbb{R}$ (or $F^\pi\in\R^{|\calX|}$ from the vector perspective) is a feasibility function of policy $\pi$ if (1) for any state $x\in\mathcal{X}$, $F^\pi(x)\leq0 \iff x\in\mathrm{X}^\pi$, and (2) there exists an operator $R^\pi:\R^{|\calX|}\to\R^{|\calX|}$ that satisfies the following three properties:
\begin{enumerate}
\item[(a)] $R^\pi F^\pi=F^\pi$.
\item[(b)] $R^\pi$ is monotonic, i.e., for any $F,F'\in\R^{|\calX|}$ such that $F\le F'$, we have $R^\pi F\le R^\pi F'$.
\item[(c)] $R^\pi$ is a $\gamma$-contraction under the infinity norm, i.e., for any $F,F'\in\R^{|\calX|}$, we have $\Vert R^\pi F-R^\pi F'\Vert_\infty\le\gamma\Vert F-F'\Vert_\infty$, where $0<\gamma<1$ is the discount factor.
\end{enumerate}
\end{definition}

The operator $R^\pi$ defined above is called the \textit{risky self-consistency operator}.
The contraction property allows us to solve $F^\pi$ by iteratively applying $R^\pi$ from an arbitrary initial function: by Banach’s fixed point theorem, for any $F\in\R^{|\calX|}$, we have
$$F^\pi=\lim_{k\to\infty}(R^\pi)^k F.$$
To better understand the above definition, we give two examples of feasibility functions.

\begin{example}[Cost value function]
\label{exp: CVF}
The cost value function (CVF) of policy $\pi$, $F^\pi_\text{CVF}$, is defined as
$$F^\pi_\text{CVF}(x_0)=\sum_{t=0}^\infty\gamma^t c(x_t),\forall x_0\in\mathcal{X},$$
where $c(x_t)=\bm{1}[h(x_t)>0]$ is the indicator function of constraint violation.
The zero-sublevel set of $F^\pi_\text{CVF}$ is the feasible region of $\pi$ because
$$
\begin{aligned}
    F^\pi_\text{CVF}(x_0)\le0 &\iff c(x_t)=0,\forall t\ge0 \\
    &\iff h(x_t)\le0,\forall t\ge0 \iff x_0\in\mathrm{X}^\pi.
\end{aligned}
$$
The risky self-consistency operator of the CVF $R^\pi_\text{CVF}$ is
$$R^\pi_\text{CVF}F^\pi_\text{CVF}(x)=c(x)+\gamma F^\pi_\text{CVF}(f(x,\pi(x))),\forall x\in\mathcal{X}.$$
The CVF is exactly in the same form as the state-value function.
Therefore, it inherits all three properties in Definition \ref{def: feasibility function} from the state-value function and is thus a feasibility function.
\end{example}

\begin{example}[Constraint decay function]
\label{exp: CDF}
The constraint decay function (CDF) of policy $\pi$, $F^\pi_\text{CDF}$, is defined as
$$F^\pi_\text{CDF}(x_0)=\gamma^{N^\pi(x_0)},\forall x_0\in\mathcal{X},$$
where $N^\pi(x_0)$ is the number of time steps that have elapsed until the first occurrence (if any) of a constraint violation starting from $x_0$ and following $\pi$.
If $x_0\in\mathrm{X}^\pi$, $N^\pi(x_0)=+\infty$, and otherwise, $N^\pi(x_0)<\infty$.
The zero-sublevel set of $F^\pi_\text{CDF}$ is the feasible region of $\pi$ because
$$
\begin{aligned}
    F^\pi_\text{CDF}(x_0)\le0 &\iff N^\pi(x_0)=+\infty \\
    &\iff h(x_t)\le0,\forall t\ge0\iff x_0\in\mathrm{X}^\pi.
\end{aligned}
$$
The risky self-consistency operator of the CDF $R^\pi_\text{CDF}$ is
$$R^\pi_\text{CDF}F^\pi_\text{CDF}(x)=c(x)+(1-c(x))\gamma F^\pi_\text{CDF}(f(x,\pi(x))),\forall x\in\mathcal{X}.$$
Yang et al.~\cite{yang2024synthesizing} have proved that the CDF satisfies the three properties in Definition \ref{def: feasibility function} and is thus a feasibility function.
\end{example}

As shown by the above two examples, the form of a feasibility function is not unique.
Any function satisfying the properties of a feasibility function can be used to solve safe RL problems.
As we will see in later sections, no matter which form of feasibility function is used, the feasible region will be identical, and the policy will also be the same within the feasible region.
This is because we only care about where the feasible states are located, not the values of the feasibility function on these states.
Therefore, without loss of generality, we make an assumption that will largely simplify the theoretical analysis of our algorithm: We assume that $F^\pi$ only takes non-negative values, i.e., $F^\pi\ge0$.
Another way to state this assumption is that for any $F^\pi$, we can construct another non-negative feasibility function $F^\pi_+=\max\{F^\pi,0\}$.
Such a constructed feasibility function will not affect the policy and the state-value function within the feasible region, and thus all theoretical analysis under this assumption also applies to general cases.

As one may notice, the risky self-consistency operator of a feasibility function $R^\pi$ is a counterpart of the \textit{self-consistency operator} of a state-value function $T^\pi:\R^{|\calX|}\to\R^{|\calX|}$, which is defined as
\begin{equation}
    T^\pi V(x)=r(x,\pi(x))+\gamma V(f(x,\pi(x))),\forall x\in\mathcal{X}.
\end{equation}
It has been proved that $T^\pi$ also satisfies the three properties in Definition \ref{def: feasibility function}.
Therefore, we can also solve $V^\pi$ by fixed point iteration:
$$V^\pi=\lim_{k\to\infty}(T^\pi)^k V,\forall V\in\R^{|\calX|}.$$

\subsection{Optimality condition for safe reinforcement learning}
The ultimate goal of safe RL is to find an optimal policy that maximizes the long-term return while having 
the maximum feasible region as its feasible region. As a result, all the analysis regarding of optimality in standard RL needs to be extended to involve both value and feasibility. Following the route of standard RL, we start by defining optimal feasibility function, optimal state-value function and optimal policy, and then establish optimality condition which describes the essential properties of them and offers a way to solve them.

Comparison between two feasibility function is direct. One feasibility function $F_1$ is considered to be better than or equal to another feasibility function $F_2$, if $F_1\le F_2$. This leads to the optimal feasibility function, which is the infimum of feasibility functions over policies. 
\begin{definition}[Optimal feasibility function]\label{def: optimal feasibility function}
$F^*: \mathcal{X}\to\mathbb{R}_{\ge0}$ is an optimal feasibility function if
\begin{equation}
    F^*=\inf_\pi F^\pi.
\end{equation}
\end{definition}
The following proposition links $F^*$ and $\rmX^*$, stating that any policy having $F^*$ as its feasibility function must also have $\rmX^*$ as its feasible region.
\begin{proposition}
\label{pro: maximum feasible region}
For any policy $\pi$ such that $F^\pi=F^*$, we have $\mathrm{X}^\pi=\mathrm{X}^*$.
\end{proposition}

With feasibility under consideration, a state-value function $V^\pi$ is no longer meaningful at every state, but only within the corresponding feasible region $\rmX^\pi$. Consequently, comparison between two state-value function is generally restricted to a certain feasible region. Besides, since feasibility should be a prior consideration to return, the optimal state-value function is defined as the supremum of state-value functions over policies with the maximum feasible region, instead of all policies.
\begin{definition}[Optimal state-value function]\label{def: Optimal state-value function}
$V^*: \mathcal{X}\to\mathbb{R}$ is an optimal state-value function if
\begin{equation}
    V^*{=_{\mathrm{X}^*}}\sup_{\pi\in\Pi^*} V^\pi,
\end{equation}
where $\Pi^*=\{\pi|\mathrm{X}^\pi=\mathrm{X}^*\}$. Note that the supremum here is based on partial order $\le_{\rmX^*}$, as can be inferred from the use of $=_{\rmX^*}$ in definition.
\end{definition}

An optimal policy in safe RL is defined on top of $F^*$ and $V^*$ as follows.
\begin{definition}[Optimal policy]
$\pi^*:\mathcal{X}\to\mathcal{U}$ is an optimal policy if
\begin{subequations}
\begin{equation}
\label{eq: optimal policy feasibility condition}
    F^{\pi^*}=F^*,
\end{equation}
\begin{equation}
\label{eq: optimal policy optimality condition}
    V^{\pi^*}{=_{\mathrm{X}^*}}V^*.
\end{equation}
\end{subequations}
\end{definition}

In order to solve the above-defined optimal functions and policy, we need also optimality version of self-consistency operators, which are termed \textit{risky Bellman operator} and \textit{feasible Bellman operator}, for feasibility functions and state-value functions correspondingly.
\begin{definition}[Risky Bellman operator]\label{def: risky bellman operator}
The risky Bellman operator $R^*:\R^{|\calX|}\to\R^{|\calX|}$ is defined as
\begin{equation}
    R^*F=\inf_\pi R^\pi F.
\end{equation}
\end{definition}

\begin{definition}[Feasible Bellman operator]\label{def: Feasible Bellman operator}
The feasible Bellman operator $T^*:\R^{|\calX|}\to\R^{|\calX|}$ is defined as
\begin{equation}
    T^*V(x)=
    \begin{cases}
        \sup_{\pi\in\Pi^*}T^\pi V(x) & x\in\mathrm{X}^*\\
        V(x) & x\notin\mathrm{X}^*
    \end{cases}
\end{equation}
\end{definition}

Until now, we have defined everything we want to solve (i.e., $F^*$, $V^*$ and $\pi^*$) and tools for solving them (i.e., $R^*$ and $T^*$). The following important theorem theoretically justifies the iterative use of risky Bellman operator and feasible Bellman operator for solving $F^*$ and $V^*$, correspondingly. It also shows the way of obtaining an optimal policy from the solved optimal functions.
\begin{theorem}[Optimality condition]
\label{thm: optimality condition}
Let $F^*$ be an optimal feasibility function and $V^*$ be an optimal state-value function, we have
% , $R^*$ be a risky Bellman operator, and $T^*$ be a feasible Bellman operator
\begin{enumerate}
\item \label{eq: condition of optimal policy}Any policy $\pi$ that satisfies
\begin{subequations}
\begin{equation}
    R^\pi F^*=R^*F^*,\label{eq: optimality cond 1}
\end{equation}
\begin{equation}
    T^\pi V^*{=_{\mathrm{X}^*}}T^*V^*,\label{eq: optimality cond 2}
\end{equation}
\end{subequations}
is an optimal policy.
\item It holds that
\begin{subequations}
\begin{equation}
\label{eq: risky Bellman equation}
    F^*=R^*F^*,
\end{equation}
\begin{equation}
\label{eq: feasible Bellman equation}
    V^*=T^*V^*.
\end{equation}
\end{subequations}
\end{enumerate}
\end{theorem}

\begin{proof}
For any policy $\pi$, we have 
\[
    \begin{array}{lr@{\;}lr}
         &F^\pi&\ge F^* &\\
        \implies& \qquad F^\pi=R^\pi F^\pi&\ge R^\pi F^*\qquad\quad&\text{(Def. \ref{def: feasibility function})}\\
        \implies& \inf_\pi F^\pi&\ge\inf_\pi R^\pi F^*& \\
        \implies& F^*&\ge R^*F^*&\text{(Def. \ref{def: optimal feasibility function}\&\ref{def: risky bellman operator})}\\
        \implies& F^*&\ge R^\pi F^*&\text{(By Eq. \eqref{eq: optimality cond 1})}\\
        \implies& F^*&\ge(R^\pi)^k F^*&\text{(Def. \ref{def: feasibility function})}\\
        \implies& F^*&\ge F^\pi&\text{(Let }k\to\infty\text{)}\\
        \implies& F^*&= F^\pi,&\text{(Def. \ref{def: optimal feasibility function})}
    \end{array}
\]
which proves the first condition \eqref{eq: optimal policy feasibility condition} of an optimal policy.
Furthermore, we have
$$F^*=F^\pi=R^\pi F^\pi=R^\pi F^*=R^*F^*,$$
which proves Eq. \eqref{eq: risky Bellman equation}.

Since $F^*=F^\pi$, we have $\mathrm{X}^\pi=\mathrm{X}^*$ by Proposition \ref{pro: maximum feasible region}, which means $\pi\in\Pi^*$. Thus, we have
\[
    \begin{array}{lr@{\;}lr}
        &V^\pi&\le\sup_{\pi\in\Pi^*}V^\pi*&\\        
        \implies&V^\pi&{\le_{\mathrm{X}^*}}V^*&\text{(Def. \ref{def: Optimal state-value function})}\\
        \implies& V^\pi=T^\pi V^\pi&{\le_{\mathrm{X}^*}}T^\pi V^*&\text{(Monotonicity)}\\
        \implies& \sup_{\pi\in\Pi^*}V^\pi&{\le_{\mathrm{X}^*}}\sup_{\pi\in\Pi^*}T^\pi V^*&\\
        \implies& V^*&{\le_{\mathrm{X}^*}}T^*V^*\quad\;\;&\text{(Def. \ref{def: Optimal state-value function}\&\ref{def: Feasible Bellman operator})}\\
        \implies& V^*&{\le_{\mathrm{X}^*}}T^\pi V^*&\text{(By Eq. \eqref{eq: optimality cond 2})}\\
        \implies& V^*&{\le_{\mathrm{X}^*}}(T^\pi)^k V^*&\text{(Monotonicity)}\\
        \implies& V^*&{\le_{\mathrm{X}^*}}V^\pi&\text{(Let }k\to\infty\text{)}\\
        \implies& V^*&{=_{\mathrm{X}^*}}V^\pi,&\text{(Def. \ref{def: Optimal state-value function})}
    \end{array}
\]
which proves the second condition \eqref{eq: optimal policy optimality condition} of an optimal policy.
Furthermore, we have
$$V^*{=_{\mathrm{X}^*}}V^\pi=T^\pi V^\pi{=_{\mathrm{X}^*}}T^\pi V^*{=_{\mathrm{X}^*}}T^*V^*,$$
which, together with $V^*{=_{\bar{\mathrm{X}}^*}}T^*V^*$, proves Eq. \eqref{eq: feasible Bellman equation} and thus finishes the proof of the theorem.
\end{proof}
Eq. \eqref{eq: risky Bellman equation} is called the \textit{risky Bellman equation}, and Eq. \eqref{eq: feasible Bellman equation} is called the \textit{feasible Bellman equation}. They indicate that, by iteratively applying $R^*$ and $T^*$, we will eventually converge to fixed points of them, which are $F^*$ and $V^*$, correspondingly. And item \ref{eq: condition of optimal policy} of Theorem \ref{thm: optimality condition} further states that, if we behave ``greedily" (in terms of both feasibility and return) according to the solved $F^*$ and $V^*$, we can obtain an optimal policy $\pi^*$. 

As suggested by their names, all the definitions and theorem introduced in this chapter together form a safe-RL extension of the optimality theory in standard RL, which is a long missing yet of significant importance piece of safe RL.

\section{Feasible Policy Iteration}
Through the above analysis, we know that the essence of safe RL is to solve the two Bellman equations \eqref{eq: risky Bellman equation} and \eqref{eq: feasible Bellman equation}.
To achieve this goal, we propose an iterative algorithm called FPI, which alternates between three steps: (1) PEV, (2) RID, and (3) region-wise PIM.
PEV is the same as in standard RL, which computes the state-value function of the current policy.
RID computes the feasibility function, which gives the feasible region of the current policy.
Region-wise PIM updates the policy to maximize its state-value function and minimize its feasibility function simultaneously.

\subsection{Policy evaluation}
In PEV, we solve the self-consistency condition of the state-value function:
\begin{equation}
\label{eq: self-consistency condition}
    V^\pi(x)=r(x,\pi(x))+\gamma V^\pi(f(x,\pi(x))),\forall x\in\mathcal{X}.
\end{equation}
As previously mentioned, this equation can be solved by iteratively applying the self-consistency operator $T^\pi$ to an arbitrary initial state-value function.
By the contraction property of $T^\pi$, we can expect geometric convergence to $V^\pi$.
However, in finite state space, we have a more straightforward method to solve $V^\pi$ because Eq. \eqref{eq: self-consistency condition} has an analytical solution.
To see this, first construct a state transition matrix $P^\pi\in\mathbb{R}^{|\mathcal{X}|\times|\mathcal{X}|}$, where $P^\pi_{ij}$ represents the probability of transferring from $x_i$ to $x_j$ under policy $\pi$.
In a deterministic MDP, each row of $P$ has a single element that equals one, which corresponds to the unique next state, and all other elements are zero.
Therefore, we have $P^\pi_{ij}=\bm{1}[f(x_i,\pi(x_i))=x_j]$.
Using this state transition matrix, and with a slight abuse of notation, we can write Eq. \eqref{eq: self-consistency condition} as
\begin{equation}
\label{eq: self-consistency condition vector}
    V^\pi=r^\pi+\gamma P^\pi V^\pi,
\end{equation}
where here we view $V^\pi$ as a vector with $|\mathcal{X}|$ elements and $r^\pi$ also a vector with elements $r^\pi(x)=r(x,\pi(x))$.
With a rearrangement of Eq. \eqref{eq: self-consistency condition vector} and noticing that the matrix $I-\gamma P^\pi$ is invertible, we can readily obtain the solution as $V^\pi=(I-\gamma P^\pi)^{-1}r^\pi$.

\subsection{Region identification}
In RID, we solve the risky self-consistency condition of the feasibility function:
\begin{equation}
\label{eq: risky self-consistency condition}
    F^\pi(x)=R^\pi F^\pi(x),\forall x\in\mathcal{X}.
\end{equation}
Similar to the self-consistency condition, Eq. \eqref{eq: risky self-consistency condition} can also be solved by iteratively applying the risky self-consistency operator $R^\pi$, which provides a geometric convergence.
Comparing Eq. \eqref{eq: risky self-consistency condition} with Eq. \eqref{eq: self-consistency condition}, a question is can we also analytically solve Eq. \eqref{eq: risky self-consistency condition}?
Without the explicit form of $R^\pi$, we cannot guarantee an analytical solution exists.
Fortunately, an analytical solution does exist for many widely-used feasibility functions.
The CVF in Example \ref{exp: CVF} obviously can be analytically solved because it has the same form as the state-value function.
As a non-trivial example, we show that the CDF in Example \ref{exp: CDF} can also be analytically solved.
The risky self-consistency condition of the CDF is
$$F^\pi(x)=c(x)+(1-c(x))\gamma F^\pi(f(x,\pi(x))),\forall x\in\mathcal{X},$$
which can be written as
$$F^\pi=c+\gamma CP^\pi F^\pi,$$
where $C\in\mathbb{R}^{|\mathcal{X}|\times|\mathcal{X}|}$ is a diagonal matrix with its elements $C_{ii}=1-c_i$.
The analytical solution of this equation can be readily obtained as $F^\pi=(I-\gamma CP^\pi)^{-1}c$.

\subsection{Region-wise policy improvement}
In region-wise PIM, we separately update the policy on states inside and outside the current feasible region.
Let $\pi$ be the original policy and $\pi'$ be the new policy.
For states outside the feasible region, we update the policy to minimize the feasibility function:
\begin{equation}
    \pi'(x)=\arg\inf_{\tilde{\pi}}R^{\tilde{\pi}}F^\pi(x),\forall x\in\bar{\mathrm{X}}^\pi.
\end{equation}
Using the risky Bellman operator, this update rule can be equivalently written as
$$R^{\pi'}F^\pi{=_{\bar{\mathrm{X}}^\pi}}R^*F^\pi.$$
Since we assume that $F^\pi$ equals zero inside the feasible region, the above equation can be simplified to $R^{\pi'}F^\pi=R^*F^\pi$.
For states inside the feasible region, we update the policy to maximize the state-value function, but the new policy should keep the current feasible region forward invariant:
\begin{equation}
\label{eq: policy update inside}
    \pi'(x)=\arg\sup_{\tilde{\pi}\in\Pi^\pi}T^{\tilde{\pi}}V^\pi(x),\forall x\in\mathrm{X}^\pi,
\end{equation}
where $\Pi^\pi$ is the set of policies that keep $\mathrm{X}^\pi$ forward invariant, i.e.,
$$\Pi^\pi=\{\tilde{\pi}|\forall x\in\mathrm{X}^\pi,f(x,\tilde{\pi}(x))\in\mathrm{X}^\pi\}.$$
To facilitate theoretical analysis of our algorithm, we introduce an operator to describe this update rule.

\begin{definition}[Restricted feasible Bellman operator]
The restricted feasible Bellman operator on the feasible region of $\pi$, $T^*_\pi:\R^{|\calX|}\to\R^{|\calX|}$ is defined as
\begin{equation}
    T^*_\pi V(x)=
    \begin{cases}
        \sup_{\tilde{\pi}\in\Pi^\pi}T^{\tilde{\pi}}V(x) & x\in\mathrm{X}^\pi,\\
        V(x) & x\notin\mathrm{X}^\pi.
    \end{cases}
\end{equation}
\end{definition}

Using the restricted feasible Bellman operator, the update rule \eqref{eq: policy update inside} can be equivalently written as
$$T^{\pi'}V^\pi{=_{\mathrm{X}^\pi}}T^*_\pi V^\pi.$$

\subsection{Theoretical analysis}
The theoretical significance of FPI lies in its two fundamental guarantees: monotonic improvement and geometric convergence, which are analogous to those in PI, but are largely extended in the context of safe RL.
Monotonic improvement refers to the consistent decrease of the feasibility function and the consistent increase of the state-value function.
Similarly, geometric convergence applies to both the feasibility and state-value functions.
Despite the theoretical foundation established by PI, extending these properties to FPI is far from straightforward.
A key challenge is that the Bellman operator is restricted to a feasible region during PIM.
This restriction ties the monotonicity of the state-value function to that of the feasible region and complicates the convergence mechanism of the state-value function.
To begin the proof, we first define the restricted optimal state-value function on a feasible region.

\begin{definition}[Restricted optimal state-value function]
$V^*_\pi:\mathcal{X}\to\mathbb{R}$ is a restricted optimal state-value function on the feasible region of $\pi$ if
\begin{equation}
    V^*_\pi{=_{\mathrm{X}^\pi}}\sup_{\pi'\in\Pi^\pi}V^{\pi'}.
\end{equation}
\end{definition}

The restricted optimal state-value function is different from, or more precisely, lower than the optimal state-value function because it maximizes over a subset of the policy space rather than the entire space.
This function represents the best achievable solution when performing PIM inside a given feasible region.
A notable special case arises when the feasible region equals the maximum one.
In this case, the restricted optimal state-value function becomes the optimal state-value function.

\begin{proposition}
For any policy $\pi$ such that $\mathrm{X}^\pi=\mathrm{X}^*$, $V^*_\pi$ is an optimal state-value function.
\end{proposition}

\begin{proof}
If we can prove $\Pi^\pi=\Pi^*$, then we have
$$V^*_\pi{=_{\mathrm{X}^*}}\sup_{\pi'\in\Pi^*}V^{\pi'},$$
which means $V^*_\pi$ is an optimal state-value function.
To prove $\Pi^\pi=\Pi^*$, we show that $\Pi^\pi\subseteq\Pi^*$ and $\Pi^*\subseteq\Pi^\pi$.
First, for any policy $\pi'\in\Pi^\pi$, we have
$$\forall x\in\mathrm{X}^\pi=\mathrm{X}^*,f(x,\pi'(x))\in\mathrm{X}^\pi=\mathrm{X}^*.$$
Thus, $\mathrm{X}^*$ is forward invariant under $\pi'$, which mean $\mathrm{X}^*\subseteq\mathrm{X}^{\pi'}$.
Together with $\mathrm{X}^{\pi'}\subseteq\mathrm{X}^*$, we have $\mathrm{X}^{\pi'}=\mathrm{X}^*$, and thus $\pi'\in\Pi^*$.
By the arbitrariness of $\pi'$, we conclude that $\Pi^\pi\subseteq\Pi^*$.
On the other hand, for any policy $\pi'\in\Pi^*$, using that $\mathrm{X}^{\pi'}$ is forward invariant under $\pi'$, we have
$$\forall x\in\mathrm{X}^{\pi'}=\mathrm{X}^*=\mathrm{X}^\pi,f(x,\pi'(x))\in\mathrm{X}^{\pi'}=\mathrm{X}^*=\mathrm{X}^\pi.$$
Thus, we have $\pi'\in\Pi^\pi$, which proves that $\Pi^*\subseteq\Pi^\pi$ and, therefore, $\Pi^\pi=\Pi^*$.
\end{proof}

This proposition reveals the dependencies of reward performance on the size of the feasible region: achieving optimal reward performance requires obtaining the maximum feasible region.
This is because the maximum feasible region provides the largest policy space for searching the optimal policy.
Just as the optimality condition for the unrestricted policy space, the restricted optimal state-value function follows a similar condition related to the restricted feasible Bellman operator.

\begin{lemma}[Restricted optimality condition]
Let $V^*_\pi$ be a restricted optimal state-value function on the feasible region of $\pi$, we have
\begin{equation}
\label{eq: restricted feasible Bellman equation}
    V^*_\pi=T^*_\pi V^*_\pi.
\end{equation}
\end{lemma}

\begin{proof}
For any policy $\pi'\in\Pi^\pi$, we have
\begin{equation}
    \begin{array}{lr@{\;}lr}
        &V^{\pi'}&\le\sup_{\tilde{\pi}\in\Pi^\pi}V^{\tilde{\pi}}{=_{\mathrm{X}^\pi}}V^*_\pi&\\        
        \implies&T^{\pi'}V^{\pi'}&{\le_{\mathrm{X}^\pi}}T^{\pi'}V^*_\pi&\text{(}\mathrm{X}^\pi\subseteq\mathrm{X}^{\pi'}\text{)}\\
        \implies&V^{\pi'}&{\le_{\mathrm{X}^\pi}}T^{\pi'}V^*_\pi&\\
        \implies&\sup_{\pi'\in\Pi^\pi}V^{\pi'}&{\le_{\mathrm{X}^\pi}}\sup_{\pi'\in\Pi^\pi}T^{\pi'}V^*_\pi&\\
        \implies&V^*_\pi&{\le_{\mathrm{X}^\pi}}T^*_\pi V^*_\pi.& \label{eq: restricted optimality proof Eq. 1}
    \end{array}
\end{equation}

Let $\pi'^*=\arg\sup_{\pi'\in\Pi^\pi}T^{\pi'}V^*_\pi$, then we have
$$T^{\pi'^*}V^*_\pi{=_{\mathrm{X}^\pi}}T^*_\pi V^*_\pi.$$
Chaining this with Eq. \eqref{eq: restricted optimality proof Eq. 1}, we have
\[
    \begin{array}{lr@{\;}lr}
        &\qquad\qquad V^*_\pi&{\le_{\mathrm{X}^\pi}}T^{\pi'^*}V^*_\pi \qquad\qquad&\\        
        \implies&V^*_\pi&{\le_{\mathrm{X}^\pi}}(T^{\pi'^*})^k V^*_\pi&\text{(Monotonicity)}\\
        \implies&V^*_\pi&{\le_{\mathrm{X}^\pi}}V^{\pi'^*}.&\text{(Let }k\to\infty\text{)}
    \end{array}
\]
Together with $V^*_\pi{\ge_{\mathrm{X}^\pi}}V^{\pi'^*}$, we have $V^*_\pi{=_{\mathrm{X}^\pi}}V^{\pi'^*}$.
Therefore, we have
$$V^*_\pi{=_{\mathrm{X}^\pi}}V^{\pi'^*}=T^{\pi'^*}V^{\pi'^*}{=_{\mathrm{X}^\pi}}T^{\pi'^*}V^*_\pi{=_{\mathrm{X}^\pi}}T^*_\pi V^*_\pi.$$
Since $T^*_\pi V^*_\pi{=_{\bar{\mathrm{X}}^\pi}}V^*_\pi$, we conclude that $V^*_\pi=T^*_\pi V^*_\pi$.
\end{proof}

Eq. \eqref{eq: restricted feasible Bellman equation} is called the \textit{restricted feasible Bellman equation}.
Through the iterations of FPI, we are essentially solving this equation to obtain the restricted optimal state-value function.
As we will later illustrate, this equation also enables us to solve the value function through an iterative process with convergence guarantees.
A key factor enabling this convergence is the monotonicity of the Bellman operators.

\begin{lemma}[Monotonicity of operators]
\label{lem: Monotonicity of operators}
For any policy $\pi$, set $\mathrm{X}\subseteq\mathrm{X}^\pi$, and functions $F,F'\in\R^{|\calX|},V,V'\in\R^{|\calX|}$ such that $F\le F',V{\le_\rmX}V'$, we have
\begin{subequations}
\begin{equation}
\label{eq: monotonicity of risky Bellman operator}
    R^*F\le R^*F',
\end{equation}
\begin{equation}
\label{eq: monotonicity of feasible Bellman operator}
    T^*_\pi V{\le_\rmX}T^*_\pi V'.
\end{equation}
\end{subequations}
\end{lemma}

\begin{proof}By monotonicity of $R^{\pi'}$, we have
\[
\begin{array}{lr@{\;}lr}
     & R^{\pi'}F&\le R^{\pi'} F'&\\
     \implies& \qquad\qquad\inf_{\pi'}R^{\pi'}F&\le\inf_{\pi'}R^{\pi'}F'\qquad&\\
     \implies& R^*F&\le R^*F',&\text{(Def. \ref{def: risky bellman operator})}
\end{array}
\]
which proves Eq. \eqref{eq: monotonicity of risky Bellman operator}.
By monotonicity of $T^{\pi'}$, we have
\[
\begin{array}{lr@{\;}lr}
     & T^{\pi'}V&{\le_\rmX}T^{\pi'}V'&\\
     \implies& \quad\;\;\sup_{\pi'\in\Pi^\pi}T^{\pi'}V&{\le_\rmX}\sup_{\pi'\in\Pi^\pi}T^{\pi'}V'&\\
     \implies& T^*_\pi V&{\le_\rmX}T^*_\pi V',&\text{(Def. \ref{def: risky bellman operator})}
\end{array}
\]
which proves Eq. \eqref{eq: monotonicity of feasible Bellman operator}.
\end{proof}

Another important property that enables the iterative solving of the Bellman equations is the contraction property.
This property is the core reason why FPI can achieve a geometric convergence speed.

\begin{lemma}[$\gamma$-contraction of Bellman operators]
For any policy $\pi$, any set $\mathrm{X}\subseteq\mathrm{X}^\pi$, and any functions $F,F'\in\R^{|\calX|},V,V'\in\R^{|\calX|}$, we have
\begin{subequations}
\begin{equation}
\label{eq: contraction of risky Bellman operator}
    \Vert R^*F-R^*F'\Vert_\infty\le\gamma\Vert F-F'\Vert_\infty,
\end{equation}
\begin{equation}
\label{eq: contraction of feasible Bellman operator}
    \Vert T^*_\pi V-T^*_\pi V'\Vert_{\mathrm{X},\infty}\le\gamma\Vert V-V'\Vert_{\mathrm{X},\infty}.
\end{equation}
\end{subequations}
\end{lemma}

\begin{proof}
We first prove that
$$\Vert\inf_\pi R^\pi F-\inf_\pi R^\pi F'\Vert_\infty\le\sup_\pi\Vert R^\pi F-R^\pi F'\Vert_\infty.$$
Let $\pi_1=\arg\inf_\pi R^\pi F,\pi_2=\arg\inf_\pi R^\pi F'$, then we have
$$
\begin{aligned}
&\Vert\inf_\pi R^\pi F-\inf_\pi R^\pi F'\Vert_\infty=\Vert R^{\pi_1} F-R^{\pi_2} F'\Vert_\infty\\
&\le\max\{\Vert R^{\pi_1} F-R^{\pi_1} F'\Vert_\infty,\Vert R^{\pi_2} F-R^{\pi_2} F'\Vert_\infty\}\\
&\le\sup_\pi\Vert R^\pi F-R^\pi F'\Vert_\infty.
\end{aligned}
$$
Therefore,
$$
\begin{aligned}
\Vert R^*F-R^*F'\Vert_\infty
&=\Vert\inf_\pi R^\pi F-\inf_\pi R^\pi F'\Vert_\infty\\
&\le\sup_\pi\Vert R^\pi F-R^\pi F'\Vert_\infty\\
&\le\gamma\Vert F-F'\Vert_\infty,
\end{aligned}
$$
which proves Eq. \eqref{eq: contraction of risky Bellman operator}. In the last inequality, we used the $\gamma$-contraction property of $R^\pi$.

With similar reasoning as above, we have
$$\Vert\sup_{\pi'\in\Pi^\pi}T^{\pi'} V-\sup_{\pi'\in\Pi^\pi}T^{\pi'}V'\Vert_{\mathrm{X},\infty}\le\sup_{\pi'\in\Pi^\pi}\Vert T^{\pi'}V-T^{\pi'}V'\Vert_{\mathrm{X},\infty}.$$
By $T^*_\pi V{=_{\mathrm{X}^\pi}}\sup_{\pi'\in\Pi^\pi}T^{\pi'} V$ and $\mathrm{X}\subseteq\mathrm{X}^\pi$, we have
$$T^*_\pi V{=_\rmX}\sup_{\pi'\in\Pi^\pi}T^{\pi'} V.$$
Therefore,
$$
\begin{aligned}
\Vert T^*_\pi V-T^*_\pi V'\Vert_{\mathrm{X},\infty}
&=\Vert\sup_{\pi'\in\Pi^\pi}T^{\pi'} V-\sup_{\pi'\in\Pi^\pi}T^{\pi'}V'\Vert_{\mathrm{X},\infty}\\
&\le\sup_{\pi'\in\Pi^\pi}\Vert T^{\pi'}V-T^{\pi'}V'\Vert_{\mathrm{X},\infty}\\
&\le\gamma\Vert V-V'\Vert_{\mathrm{X},\infty},
\end{aligned}
$$
which proves Eq. \eqref{eq: contraction of feasible Bellman operator}. In the last inequality, we used the $\gamma$-contraction property of $T^{\pi'}$.
\end{proof}

With the above lemmas regarding the Bellman operators, we are ready to introduce the first key theoretical property of FPI: monotonic improvement.
This property characterizes the relationship between two policies generated in successive iterations of FPI.

\begin{theorem}[Monotonic improvement]
\label{thm: Monotonic improvement}
Let $\pi,\pi'$ be two policies such that
\begin{subequations}
\begin{equation}
    R^{\pi'}F^\pi=R^*F^\pi,
\end{equation}
\begin{equation}
    T^{\pi'}V^\pi{=_{\mathrm{X}^\pi}}T^*_\pi V^\pi.
\end{equation}
\end{subequations}
Then, we have
\begin{subequations}
\begin{equation}
\label{eq: monotonicity of feasibility function}
    F^\pi\ge R^*F^\pi\ge F^{\pi'},
\end{equation}
\begin{equation}
\label{eq: monotonicity of state-value function}
    V^\pi\le T^*_\pi V^\pi{\le_{\mathrm{X}^\pi}}V^{\pi'}.
\end{equation}
\end{subequations}
\end{theorem}

\begin{proof}
By $F^\pi=R^\pi F^\pi\ge R^*F^\pi$, we have
\begin{equation}
\label{eq: monotonic proof feasibility condition}
    F^\pi\ge R^*F^\pi=R^{\pi'}F^\pi.
\end{equation}
We prove by induction that for $k\ge1$,
\begin{equation}
\label{eq: monotonic proof feasibility induction}
    F^\pi\ge R^*F^\pi\ge(R^{\pi'})^kF^\pi.
\end{equation}
From this, Eq. \eqref{eq: monotonicity of feasibility function} will follow by taking $k\to\infty$ on both sides.
The case of $k=1$ has already been proved.
Assume that the required inequality holds for $k\ge1$, and we prove that it also holds for $k+1$.
Applying $R^{\pi'}$ on both sides of Eq. \eqref{eq: monotonic proof feasibility induction} and using monotonicity of $R^{\pi'}$, we have
$$R^{\pi'}F^\pi\ge(R^{\pi'})^{k+1}F^\pi.$$
Chaining this with Eq. \eqref{eq: monotonic proof feasibility condition}, we have
$$F^\pi\ge R^*F^\pi=R^{\pi'}F^\pi\ge(R^{\pi'})^{k+1}F^\pi,$$
which finishes the inductive step, and thus the proof of Eq. \eqref{eq: monotonic proof feasibility induction}.

By $V^\pi=T^\pi V^\pi\le T^*_\pi V^\pi$, we have
\begin{equation}
\label{eq: monotonic proof state-value condition}
    V^\pi\le T^*_\pi V^\pi{=_{\mathrm{X}^\pi}}T^{\pi'}V^\pi.
\end{equation}
We prove by induction that for $k\ge1$,
\begin{equation}
\label{eq: monotonic proof state-value induction}
    V^\pi\le T^*_\pi V^\pi{\le_{\mathrm{X}^\pi}}(T^{\pi'})^k V^\pi.
\end{equation}
From this, Eq. \eqref{eq: monotonicity of state-value function} will follow by taking $k\to\infty$ on both sides.
The case of $k=1$ has already been proved.
Assume that the required inequality holds for $k\ge1$, and we prove that it also holds for $k+1$.
Applying $T^{\pi'}$ on both sides of Eq. \eqref{eq: monotonic proof state-value induction} and using monotonicity of $T^{\pi'}$, we have
$$T^{\pi'}V^\pi{\le_{\mathrm{X}^\pi}}(T^{\pi'})^{k+1}V^\pi.$$
Chaining this with Eq. \eqref{eq: monotonic proof state-value condition}, we have
$$V^\pi\le T^*_\pi V^\pi{=_{\mathrm{X}^\pi}}T^{\pi'}V^\pi{\le_{\mathrm{X}^\pi}}(T^{\pi'})^{k+1}V^\pi,$$
which finishes the inductive step, and thus the proof of Eq. \eqref{eq: monotonic proof state-value induction}.
\end{proof}

Monotonic improvement ensures that the policy generated in successive iterations of FPI surpasses its predecessors in terms of both safety and performance.
This property enables FPI to avoid the oscillation problem commonly observed in other methods, such as the Lagrange method, and as we will see in our experiments, this leads to stable training processes.
A direct proposition of the monotonic decrease of the feasibility function is that the feasible region monotonically expands.

\begin{proposition}[Monotonic region expansion]
Let $\pi,\pi'$ be two policies such that $R^{\pi'}F^\pi=R^*F^\pi$, we have $\mathrm{X}^\pi\subseteq\mathrm{X}^{\pi'}$.
\end{proposition}

\begin{proof}
By Theorem \ref{thm: Monotonic improvement}, we have $F^\pi\ge F^{\pi'}$. Thus, for any $x\in\mathrm{X}^\pi$, we have $F^{\pi'}(x)\le F^\pi(x)\le0$, which means $x\in\mathrm{X}^{\pi'}$. Therefore, $\mathrm{X}^\pi\subseteq\mathrm{X}^{\pi'}$.
\end{proof}

Before proving geometric convergence, we need to answer another question: when optimizing the state-value function within different feasible regions, do larger regions always yield higher values?
This question is crucial because the feasible regions in each iteration of FPI are different, but the monotonicity of the state-value function we just proved is based on a fixed region.
To extend the monotonicity to these changing regions, we need to show that optimizing over a sequence of expanding regions still preserves monotonicity, which is stated by the following lemma.

\begin{lemma}[Feasible region superiority]
\label{lem: Feasible region superiority}
Let $\pi,\pi'$ be two policies such that $\mathrm{X}^\pi\subseteq\mathrm{X}^{\pi'}$, we have
\begin{equation}
    T^*_\pi V{\le_{\mathrm{X}^\pi}}T^*_{\pi'}V.
\end{equation}
\end{lemma}

\begin{proof}
We first prove that for any policy $\tilde{\pi}\in\Pi^\pi$, there exists a policy $\tilde{\pi}'\in\Pi^{\pi'}$ such that $\tilde{\pi}'{=_{\mathrm{X}^\pi}}\tilde{\pi}$.
Such a policy can be constructed as
$$\tilde{\pi}'(x)=\begin{cases}
    \tilde{\pi}(x) & x\in\mathrm{X}^\pi,\\
    \pi'(x) & x\notin\mathrm{X}^\pi,
\end{cases}$$
which naturally satisfies $\tilde{\pi}'{=_{\mathrm{X}^\pi}}\tilde{\pi}$, and the left is to prove $\tilde{\pi}'\in\Pi^{\pi'}$.
To this end, we prove that for any $x\in\mathrm{X}^{\pi'}$, we have $f(x,\tilde{\pi}'(x))\in\mathrm{X}^{\pi'}$ by considering two cases: (1) $x\in\mathrm{X}^\pi$, and (2) $x\in\mathrm{X}^{\pi'}\setminus\mathrm{X}^\pi$.
If $x\in\mathrm{X}^\pi$, we have
$$f(x,\tilde{\pi}'(x))=f(x,\tilde{\pi}(x))\in\mathrm{X}^\pi\subseteq\mathrm{X}^{\pi'}.$$
If $x\in\mathrm{X}^{\pi'}\setminus\mathrm{X}^\pi$, we have
$$f(x,\tilde{\pi}'(x))=f(x,\pi'(x))\in\mathrm{X}^{\pi'}.$$
Thus, we conclude that $\tilde{\pi}'\in\Pi^{\pi'}$.
Let $\tilde{\pi}^*=\arg\sup_{\tilde{\pi}\in\Pi^\pi}T^{\tilde{\pi}}V$.
By the conclusion above, there exists $\tilde{\pi}'^*\in\Pi^{\pi'}$ such that $\tilde{\pi}'^*{=_{\mathrm{X}^\pi}}\tilde{\pi}^*$.
Thus, we have
$$
\begin{aligned}
    T^*_\pi V &{=_{\mathrm{X}^\pi}} \sup_{\tilde{\pi}\in\Pi^\pi}T^{\tilde{\pi}}V {=_{\mathrm{X}^\pi}} T^{\tilde{\pi}^*}V \\
    &{=_{\mathrm{X}^\pi}} T^{\tilde{\pi}'^*}V{\le_{\mathrm{X}^\pi}}\sup_{\tilde{\pi}'\in\Pi^{\pi'}}T^{\tilde{\pi}'}V {=_{\mathrm{X}^\pi}} T^*_{\pi'}V,
\end{aligned}
$$
where the last equality uses that $\mathrm{X}^\pi\subseteq\mathrm{X}^{\pi'}$.
\end{proof}

Now we have everything needed for proving the most important property of FPI: geometric convergence.
This property includes convergence of the feasibility function to the global optimum and convergence of the state-value function to the restricted optimum on the feasible regions.

\begin{theorem}[Geometric convergence]
\label{thm: Geometric convergence}
Let $\{\pi_k\}_{k\ge0}$ be the sequence of policies produced by FPI.
Then, for any $k\ge0$,
\begin{subequations}
\begin{equation}
\label{eq: convergence of feasibility function}
    \Vert F^*-F^{\pi_k}\Vert_\infty\le\gamma^k\Vert F^*-F^{\pi_0}\Vert_\infty,
\end{equation}
\begin{equation}
\label{eq: convergence of state-value function}
    \Vert(V^*_{\pi_0}-V^{\pi_k})_+\Vert_{\mathrm{X}^{\pi_0},\infty}\le\gamma^k\Vert V^*_{\pi_0}-V^{\pi_0}\Vert_{\mathrm{X}^{\pi_0},\infty}.
\end{equation}
\end{subequations}
\end{theorem}

\begin{proof}
By Theorem \ref{thm: Monotonic improvement}, we have
\begin{subequations}
\begin{equation}
\label{eq: convergence proof feasibility condition}
    R^*F^{\pi_0}\ge F^{\pi_1},
\end{equation}
\begin{equation}
\label{eq: convergence proof state-value condition}
    T^*_{\pi_0}V^{\pi_0}{\le_{\mathrm{X}^{\pi_0}}}V^{\pi_1}.
\end{equation}
\end{subequations}
Applying $R^*$ to both sides of Eq. \eqref{eq: convergence proof feasibility condition} and using the monotonicity of $R^*$, we have
$$(R^*)^2 F^{\pi_0}\ge R^*F^{\pi_1}\ge F^{\pi_2},$$
where the second inequality is obtained by using Theorem \ref{thm: Monotonic improvement} again.
With the same reasoning, we have
$$(R^*)^k F^{\pi_0}\ge F^{\pi_k}.$$
Thus,
$$0\le F^{\pi_k}-F^*\le(R^*)^k F^{\pi_0}-F^*.$$
Taking the infinity norm, we have
$$
\begin{aligned}
    \Vert F^{\pi_k}-F^*\Vert_\infty
    &\le\Vert(R^*)^k F^{\pi_0}-F^*\Vert_\infty\\
    &=\Vert(R^*)^k F^{\pi_0}-(R^*)^k F^*\Vert_\infty\\
    &\le\gamma^k\Vert F^{\pi_0}-F^*\Vert_\infty,
\end{aligned}
$$
which proves Eq. \eqref{eq: convergence of feasibility function}.
In the equality above we used the optimality condition and in the last inequality we used the contraction property.

Applying $T^*_{\pi_1}$ to both sides of Eq. \eqref{eq: convergence proof state-value condition} and using the monotonicity of $T^*_{\pi_1}$, we have
$$T^*_{\pi_1}T^*_{\pi_0}V^{\pi_0}{\le_{\mathrm{X}^{\pi_0}}}T^*_{\pi_1} V^{\pi_1}{\le_{\mathrm{X}^{\pi_1}}}V^{\pi_2}.$$
By Lemma \ref{lem: Feasible region superiority} and $\mathrm{X}^{\pi_0}\subseteq\mathrm{X}^{\pi_1}$, we have
$$(T^*_{\pi_0})^2 V^{\pi_0}{\le_{\mathrm{X}^{\pi_0}}}T^*_{\pi_1}T^*_{\pi_0}V^{\pi_0}{\le_{\mathrm{X}^{\pi_0}}}V^{\pi_2}.$$
With the same reasoning, we have
$$(T^*_{\pi_0})^k V^{\pi_0}{\le_{\mathrm{X}^{\pi_0}}}V^{\pi_k}.$$
Thus,
$$V^*_{\pi_0}-V^{\pi_k}{\le_{\mathrm{X}^{\pi_0}}}V^*_{\pi_0}-(T^*_{\pi_0})^k V^{\pi_0}.$$
Applying $(\cdot)_+$ on the left hand side and taking the absolute value of the right hand side, we have
$$(V^*_{\pi_0}-V^{\pi_k})_+{\le_{\mathrm{X}^{\pi_0}}}|V^*_{\pi_0}-(T^*_{\pi_0})^k V^{\pi_0}|.$$
Taking the infinity norm on both sides, we have
$$
\begin{aligned}
    \Vert(V^*_{\pi_0}-V^{\pi_k})_+\Vert_{\mathrm{X}^{\pi_0},\infty}
    &\le\Vert V^*_{\pi_0}-(T^*_{\pi_0})^k V^{\pi_0}\Vert_{\mathrm{X}^{\pi_0},\infty}\\
    &=\Vert(T^*_{\pi_0})^k V^*_{\pi_0}-(T^*_{\pi_0})^k V^{\pi_0}\Vert_{\mathrm{X}^{\pi_0},\infty}\\
    &\le\gamma^k\Vert V^*_{\pi_0}-V^{\pi_0}\Vert_{\mathrm{X}^{\pi_0},\infty},
\end{aligned}
$$
where the equality uses the restricted optimality condition.
This proves Eq. \eqref{eq: convergence of state-value function} and thus finished the proof.
\end{proof}

As stated in this theorem, the optimal state-value function is restricted to the feasible region of the initial policy.
While this may seem overly restrictive, it is important to note that any policy in the sequence $\{\pi_k\}_{k\ge0}$ can be viewed as an initial policy.
This is because any subsequence of policies is still generated by FPI and retains the convergence property. 
Given the geometric convergence of the feasibility function, we can expect that the feasible region to quickly approach the maximum feasible region.
When the feasible region grows to a satisfying size, we can analyze the policy sequence from this point onward, leveraging the convergence of the state-value function.
In this case, the state-value function continues to exhibit geometric convergence to the optimum restricted to this feasible region, which is already very close to the global optimum.

In MDPs with finite state and action spaces, FPI achieves convergence within a finite number of iterations.
This is because the total number of deterministic policies is finite, and both the feasibility function and state-value function improve monotonically.
If two policies share identical feasibility functions, the maximum feasible region has been reached.
Subsequently, if the state-value functions of any two successive policies are identical within this region, the optimal state-value function has been attained.
However, this analysis is naive because it first requires the feasibility function to converge before turning to the state-value function.
This is impractical because of the low computational efficiency.
The significance of Theorem \ref{thm: Geometric convergence} is to show that state-value computation can start at any point, even at the very beginning, instead of waiting for the feasibility function to converge.
Regardless of the starting point, the state-value function always exhibits geometric convergence to the optimal value, restricted to a feasible region that itself converges to the maximum feasible region. 

\section{Practical implementations}
Our theoretical analysis of FPI is based on finite state and action spaces, where we can use tables to store values of feasibility function, state-value function, and policy and update them by exact value assignments.
However, when it comes to infinite state and action spaces, tabular methods no longer apply, and function approximation becomes necessary.
Similar to standard deep RL algorithms, we approximate the feasibility function, state-value function, and policy using neural networks and update them via gradient descent.
% Moreover, techniques in deep RL algorithms, such as double Q networks and maximum entropy RL, can also be applied to our FPI algorithm with minimal adaptation.
In this section, we introduce some practical implementations for function approximation used in our experiments.
% It is worth mentioning that these implementations are not the only practical methods, and there are probably other methods that are also practical and even have better performance, which are left for future research. 

On the basis of FPI, we adopt SAC~\cite{haarnoja2018soft} for reward optimization and denote the resulting algorithm as FPI-SAC.
Our algorithm learns an action feasibility function $G_\phi(x,u)$ represented by a neural network with parameter $\phi$, which is a Q-function-like variant of the CDF in Example \ref{exp: CDF}.
It takes the current state and action as input and outputs the CDF value of the next state, i.e., $G^\pi(x,u)=F^\pi(f(x,u))$.
FPI-SAC also learns two action-value networks $Q_{\omega_1}, Q_{\omega_2}$ and a stochastic policy network $\pi_\theta$.
% The policy is stochastic and outputs the mean and variance of a diagonal Gaussian distribution as in standard SAC.
% All networks have two hidden layers with 256 neurons each and use the ReLU activation function.

We use a sigmoid output activation function for $G_\phi$ because its value is within $[0,1]$ by definition.
Incorporating this prior structure into the network makes it easier to converge because it eliminates excessively large errors introduced by bootstrapping.
% A problem with the sigmoid activation function is that it makes the gradient small when the absolute value of the input is large, which slows down the training process.
To avoid the gradient vanishing problem caused by sigmoid function, we use a cross-entropy loss to train $G_\phi$, with the right-hand side of \eqref{eq: risky self-consistency condition} as the training label:
% which applies a logarithm function to the output and magnifies the gradient farther from the origin.
% Using the right-hand side of the risky self-consistency condition \eqref{eq: risky self-consistency condition} as the training label, the loss function $G_\phi$ can be written as
\begin{equation}
\begin{aligned}
    L_G(\phi)=-\mathbb{E}_{(x,u)\sim\mathcal{D}}\{y_G(x)\log G_\phi(x,u) \\
    +(1-y_G(x))\log(1-G_\phi(x,u))\}, \\
    y_G(x)=c(x)+(1-c(x))\gamma G_{\bar{\phi}}(x',u'),
\end{aligned}
\end{equation}
where $\mathcal{D}$ is the replay buffer and $\bar{\phi}$ is the parameters of the target network for $G_\phi$, which is updated at a slower rate to stabilize the training process.

The loss functions of the Q networks are the same as those in SAC:
\begin{equation}
\begin{aligned}
    L_Q(\omega_i)&=\mathbb{E}_{(x,u,r,x')\sim\mathcal{D}}\{(y_Q-Q_{\omega_i}(x,u))^2\}, \\
    y_Q&=r+\gamma(\min_{j\in\{1,2\}} Q_{\bar{\omega}_j}(x',u')-\alpha\log\pi_\theta(u'|x')),
\end{aligned}
\end{equation}
where $i\in\{1,2\}$, $\bar{\omega}_j$ are the parameters of the target Q networks, and $\alpha$ is the temperature.

The loss function of the policy is computed separately inside and outside the feasible region.
% In theory, the feasible region is represented by the zero-sublevel set of $G_\phi$.
% However, due to our usage of the sigmoid activation function, the value of $G_\phi$ is within $(0,1)$.
% Therefore, it does not have a zero-sublevel set to represent the feasible region.
% To deal with this problem, 
We introduce a feasibility threshold $0<p<1$, and use $\mathrm{X}^F=\{x\in\mathcal{X}|F_\phi(x)<p\}$ as an approximate feasible region.
A case where this approximation is valid is that constraint violations can be avoided in an infinite horizon as long as they can be avoided in a finite number of steps~\cite{thomas2021safe}.
% For a given threshold $p$, this finite number of steps is $\log_\gamma p$, and $G_\phi$ must be greater than $p$ on any infeasible state.
In our experiments, we found that a constant value $p=0.1$ can achieve zero constraint violation on all classic control tasks.
Inside the feasible region, we update the policy according to Eq. \eqref{eq: policy update inside}.
% , which maximizes the action-value function under the constraint of the feasibility function.
Since the policy from the previous iteration is feasible inside its own feasible region, it can be viewed as a feasible initial point.
This allows us to use the interior point method to solve the optimization problem in Eq. \eqref{eq: policy update inside}.
We apply a logarithm function to the constraint and add it to the objective function, obtaining the following policy loss:
\begin{equation}
\label{eq: policy loss inside feasible region}
\begin{aligned}
    L_{\pi,\text{f}}(\theta)=\mathbb{E}_{x\sim\mathcal{D},u\sim\pi_\theta(\cdot|x)}\{\bm{1}[G_\phi(x,u)<p](\alpha\log\pi_\theta(u|x)- \\
    \min_{i\in\{1,2\}}Q_{\omega_i}(x,u)-1/t\cdot\log(p-G_\phi(x,u)))\}.
\end{aligned}
\end{equation}
We increase the weight $t$ by a factor every several iterations, which is a standard practice in the interior point method.
As $t\to\infty$, the minima of Eq. \eqref{eq: policy loss inside feasible region} approaches the solution of Eq. \eqref{eq: policy update inside}.
Compared with other constrained optimization methods such as the dual ascent, the advantage of the interior point method is that its intermediate solutions are always feasible regardless of the value of $t$.
This ensures that the monotonic expansion property of the feasible region is preserved.
% In contrast, the intermediate solutions in dual ascent is not guaranteed to be feasible, but is affected by the value of the Lagrange multiplier.
% It is difficult to find an appropriate multiplier that ensures strict feasibility, resulting in an oscillating training process~\cite{stooke2020responsive,peng2022model}.
Outside the feasible region, we minimize the action feasibility function:
\begin{equation}
\label{eq: policy loss outside feasible region}
    L_{\pi,\text{i}}(\theta)=\mathbb{E}_{x\sim\mathcal{D},u\sim\pi_\theta(\cdot|x)}\{G_\phi(x,u)\}.
\end{equation}
With \eqref{eq: policy loss inside feasible region} and \eqref{eq: policy loss outside feasible region}, the total policy loss is
\begin{equation}
    L_\pi(\theta)=L_{\pi,\text{f}}(\theta)+L_{\pi,\text{i}}(\theta).
\end{equation}
The loss function of the temperature $\alpha$ is the same as that in SAC:
\begin{equation}
    L(\alpha)=\mathbb{E}_{x\sim D}\left\{-\alpha\log\pi_\theta(u|x)-\alpha\bar{\mathcal{H}}\right\},
\end{equation}
where $\bar{\mathcal{H}}$ is the target entropy.

\section{Experiments}
We aim to answer the following two questions through our experiments:
\begin{enumerate}
    \item Does FPI outperform existing safe RL algorithms in terms of safety and optimality?\label{question 1}
    \item Does FPI achieve monotonic feasible region expansion and monotonic value function improvement, and eventually converge to the maximum feasible region and optimal value function?\label{question 2}
\end{enumerate}

To answer Question \ref{question 1}, we perform extensive experiments on twelve safe RL environments, including grid world tasks, classic control tasks, and high-dimensional robot navigation tasks.
% , where FPI is compared with mainstream safe RL algorithms.
To answer Question \ref{question 2}, we visualize the feasible region and state-value function during training process to check monotonicity, and we also compare the final feasible regions with the maximum ones and the final reward performance with that of an optimal controller to check optimality.

\subsection{Environments}
\label{subsec: environments}
\begin{figure}[htbp]
    \centering
    \subfloat[SafeInterruptibility]{\includegraphics[width=0.292\linewidth]{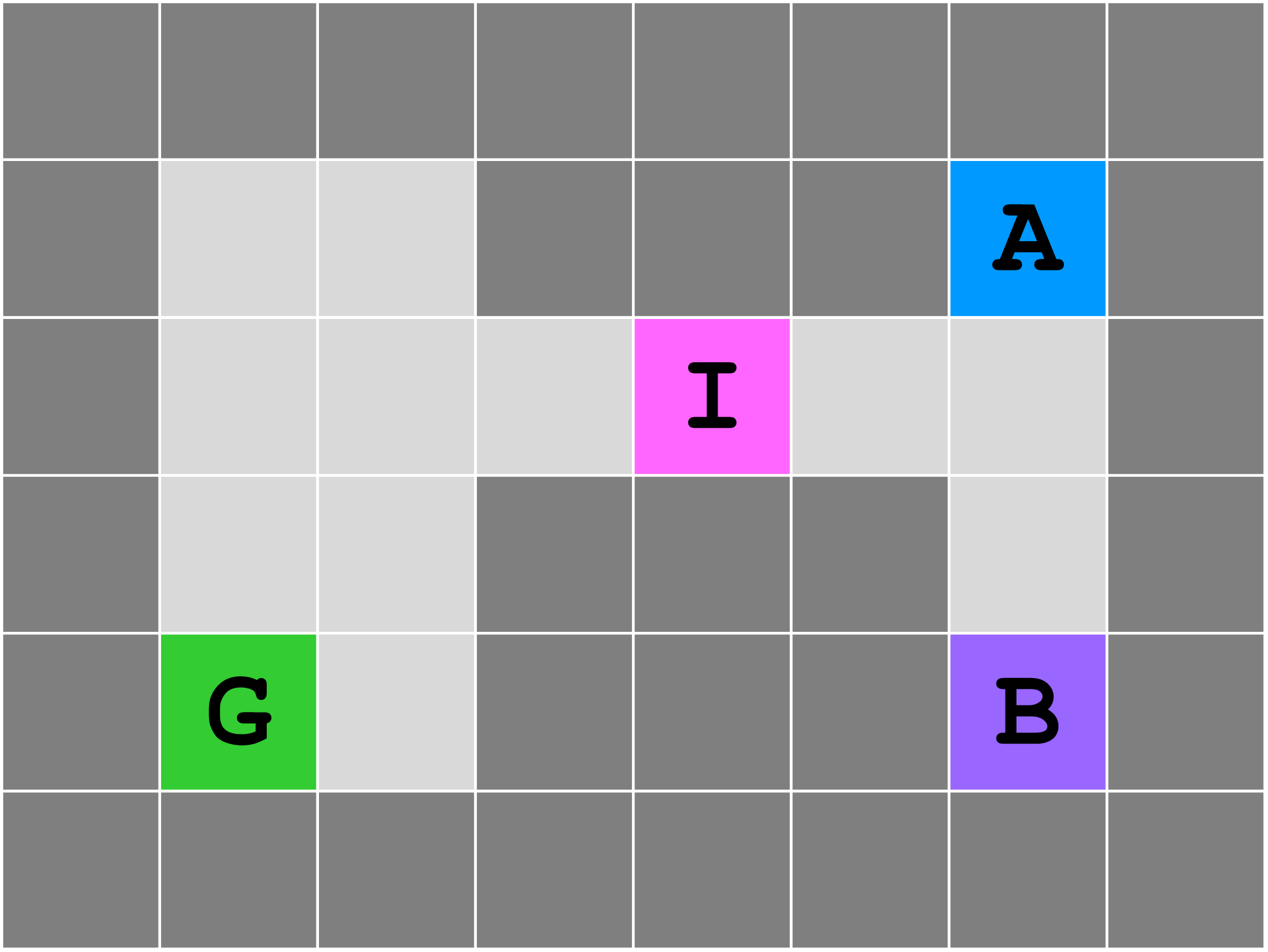}}
    \hspace{0.5pt}
    \subfloat[SideEffects]{\includegraphics[width=0.22\linewidth]{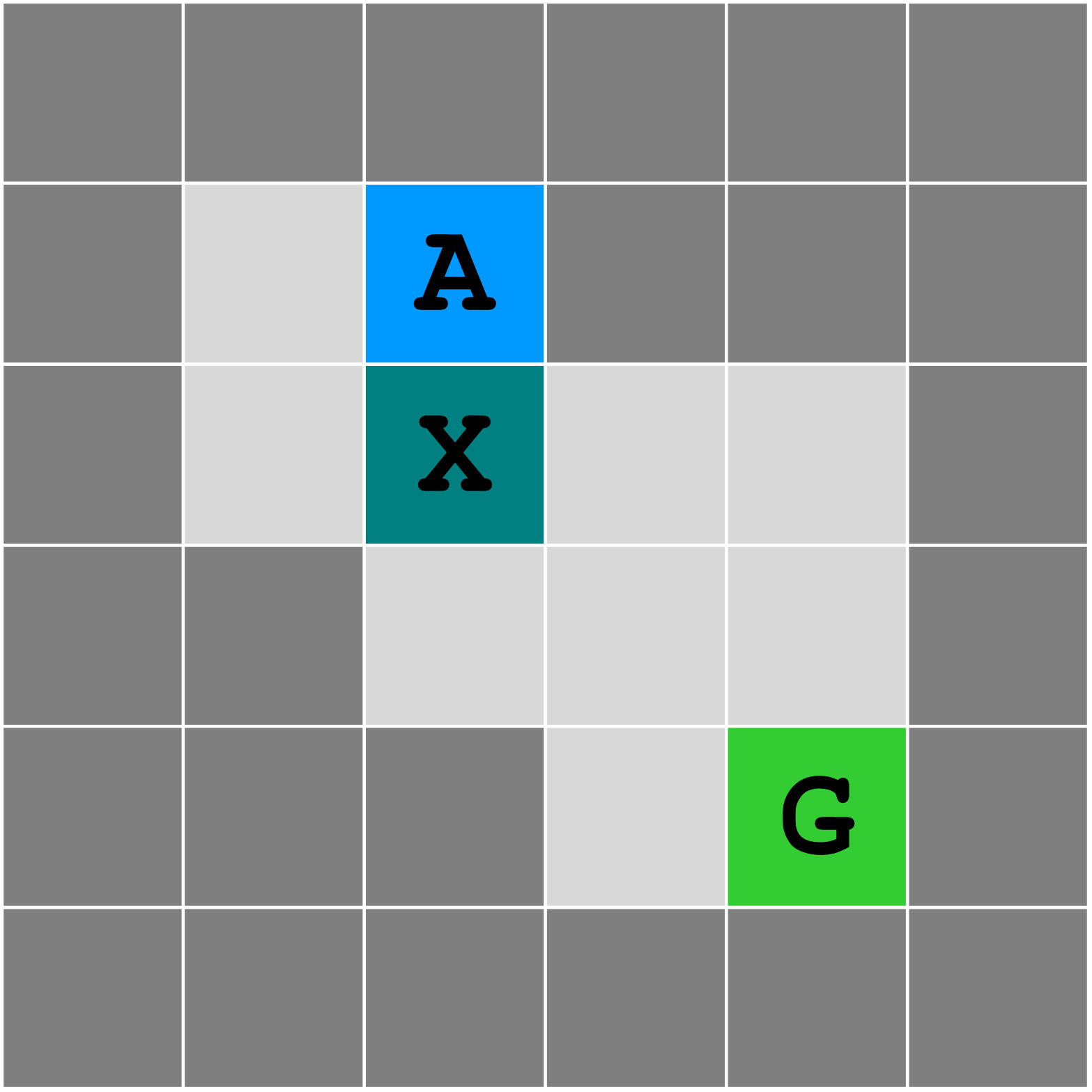}}
    \hspace{0.5pt}
    \subfloat[Supervisor]{\includegraphics[width=0.22\linewidth]{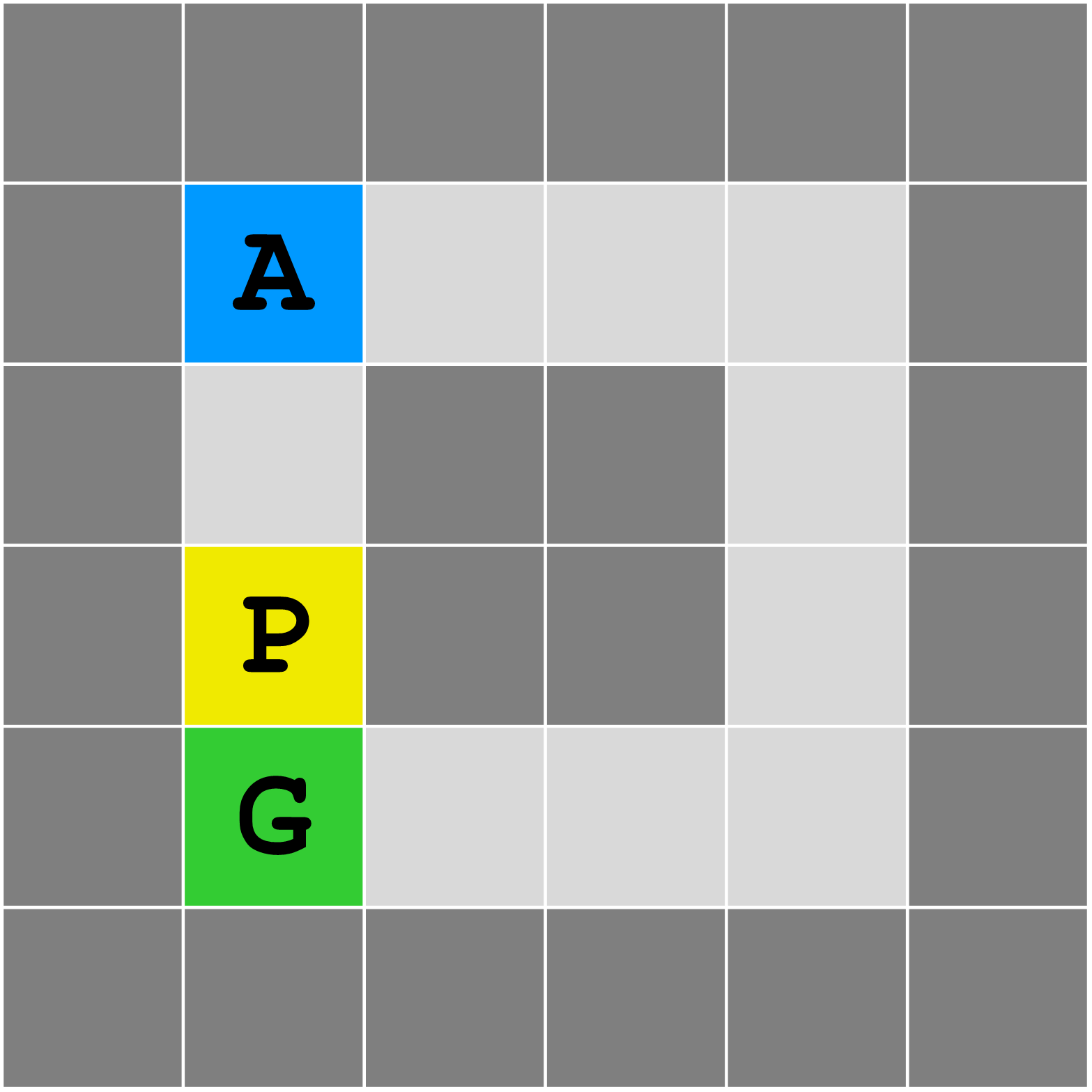}}
    \hspace{0.5pt}
    \subfloat[BoatRace]{\includegraphics[width=0.22\linewidth]{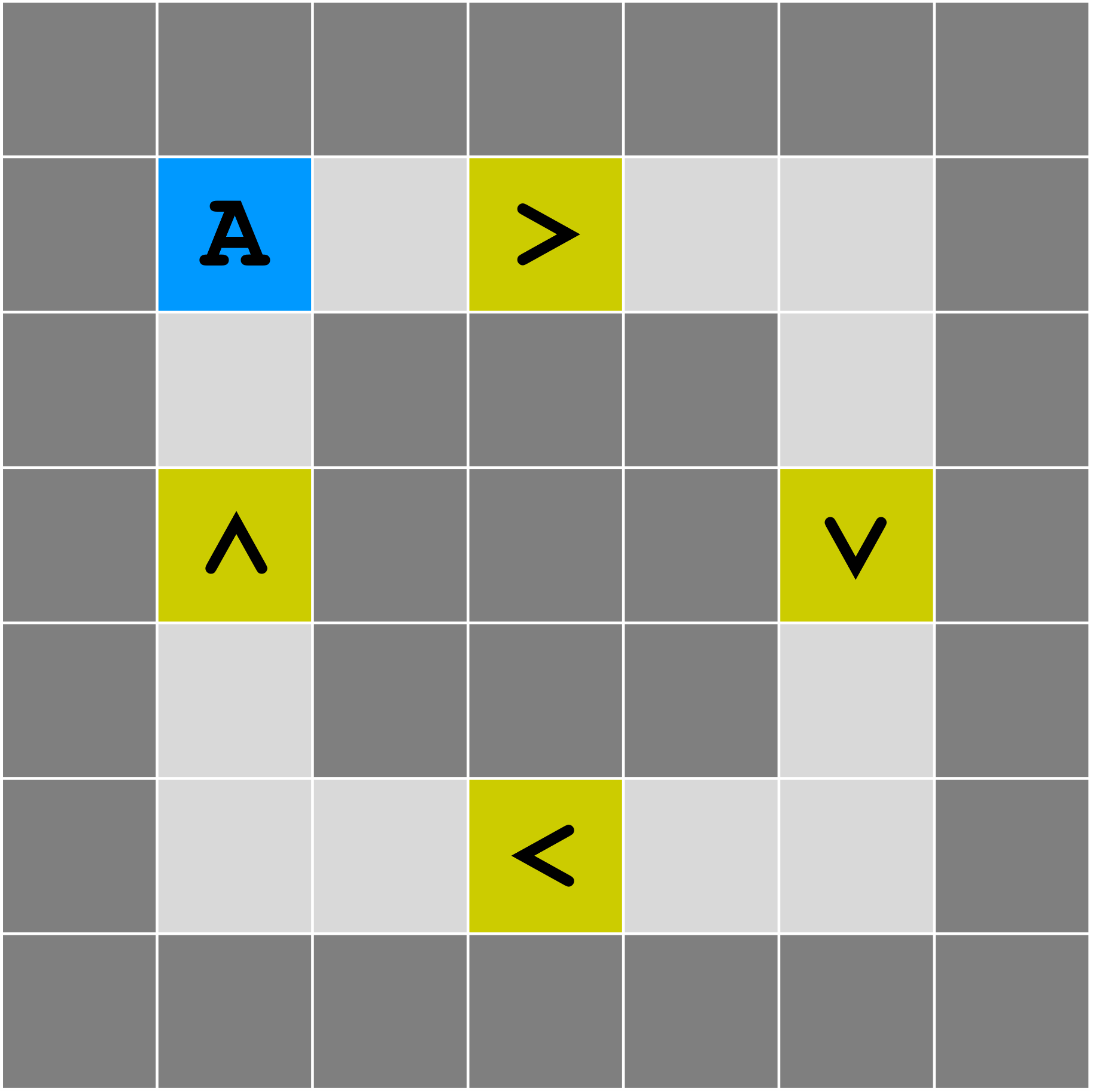}}
    \caption{Four grid world tasks. \textbf{\texttt{A}} denotes the agent. \textbf{\texttt{G}} denotes the goal. \textbf{\texttt{I}} in (a) denotes the interruption. \textbf{\texttt{B}} in (a) denotes the button. \textbf{\texttt{X}} in (b) denotes the box. \textbf{\texttt{P}} in (c) denotes the punishment. The arrows in (d) denote the checkpoints.}
    \label{fig: grid world}
\end{figure}

We choose four grid world tasks, four classic control tasks, and four high-dimensional robot navigation tasks.
The grid world tasks are borrowed from AI Safety Gridworlds~\cite{leike2017ai} with slight modifications, as shown in Fig. \ref{fig: grid world}.
In \textbf{SafeInterruptibility}, the agent should first press the button to disable the interruption, which is defined as a constraint, and then pass through to reach the goal.
In \textbf{SideEffects}, the agent should push the box to clear a path to the goal. The constraint is that the box should not be pushed to a corner.
In \textbf{Supervisor}, the agent receives a negative reward for stepping on the punishment with a 50\% probability. The constraint requires the agent to always avoid the punishment.
In \textbf{BoatRace}, the agent is rewarded for moving clockwise through the checkpoints. However, revisiting the same checkpoint back and forth results in constraint violations.
For more details about the tasks, please refer to the original paper~\cite{leike2017ai}.

\begin{figure}[htbp]
    \centering
    \subfloat[ACC]{\includegraphics[width=0.3\linewidth, trim=23 15 48 20, clip]{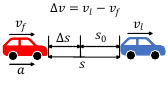}}
    \hspace{2pt}
    \subfloat[LK]{\includegraphics[width=0.22\linewidth, trim=20 60 40 15, clip]{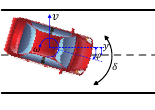}}
    \hspace{0.5pt}
    \subfloat[Pendulum]{\includegraphics[width=0.18\linewidth, trim=-30 0 -30 40, clip]{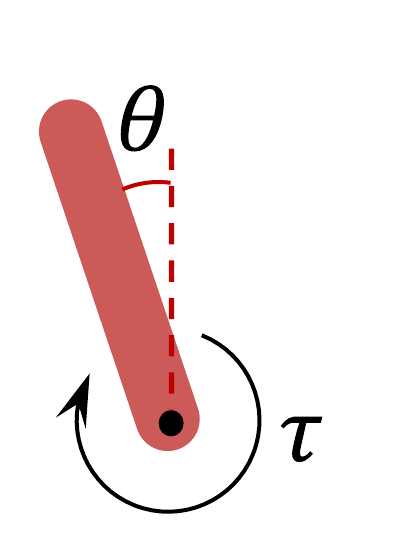}}
    \hspace{0.5pt}
    \subfloat[Quadrotor]{\includegraphics[width=0.23\linewidth, trim=0 0 0 10, clip]{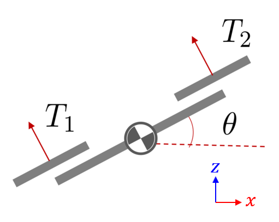}}
    \caption{Schematics of four classic control tasks.}
    \label{fig: classic control task}
\end{figure}

The classic control tasks include regulation and trajectory tracking, covering both linear and nonlinear systems, as shown in Fig. \ref{fig: classic control task}.
\textbf{ACC} requires controlling a vehicle to follow a leading vehicle while keeping the distance within given limits.
\textbf{LK} requires keeping a vehicle in a lane by controlling its steering angle. 
\textbf{Pendulum} requires swinging a pendulum to the upright position while keeping its angular position within given limits. 
\textbf{Quadrotor} is borrowed from safe-control-gym~\cite{yuan2022safe}, where a 2D quadrotor is required to follow a circular trajectory in the vertical plane while keeping the vertical position with given limits.
We defer the details of these environments, including the design of their state spaces, action spaces, dynamics, reward functions, and constraints, to Appendix \ref{sec: appx-env}.

\begin{figure}[htbp]
    \centering
    \subfloat[PointGoal]{\includegraphics[width=0.23\linewidth, trim=60 70 60 100, clip]{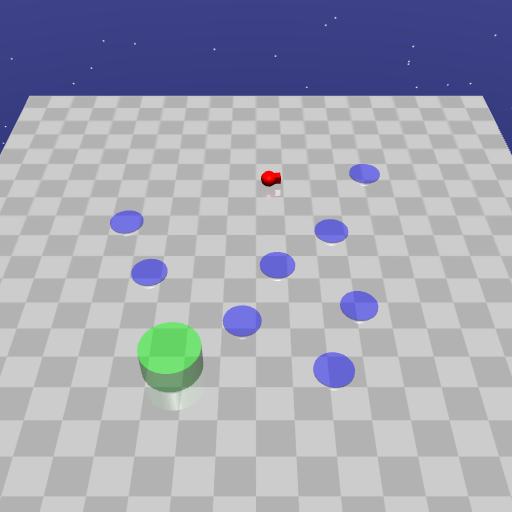}}
    \hspace{0.5pt}
    \subfloat[CarGoal]{\includegraphics[width=0.23\linewidth, trim=60 80 60 90, clip]{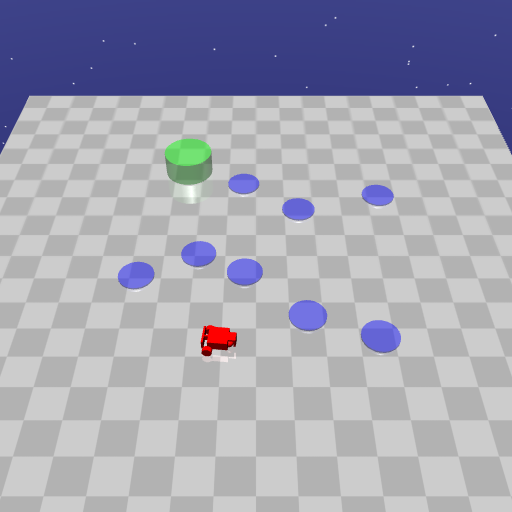}}
    \hspace{0.5pt}
    \subfloat[PointPush]{\includegraphics[width=0.23\linewidth, trim=50 80 70 90, clip]{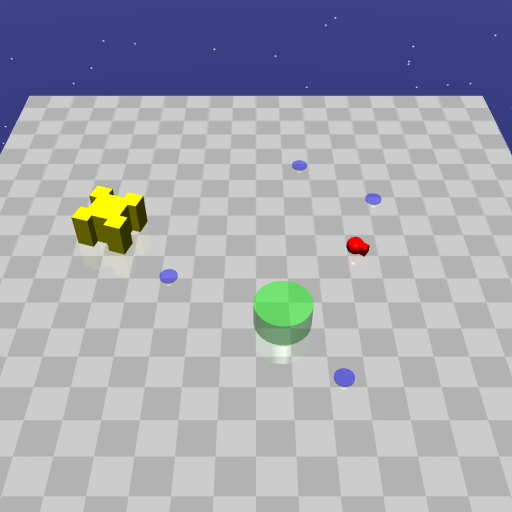}}
    \hspace{0.5pt}
    \subfloat[CarPush]{\includegraphics[width=0.23\linewidth, trim=60 80 60 90, clip]{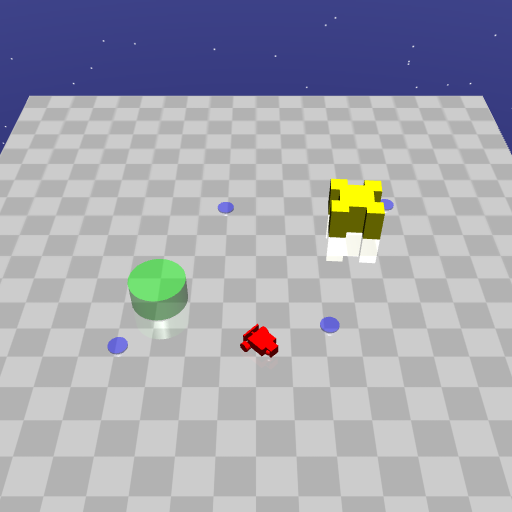}}
    \caption{Snapshots of four robot navigation tasks. The robots are in red. The blue circles represent the hazards. The Green cylinders represent the goals. The yellow objects represent the boxes.}
    \label{fig: safety gym snapshots}
\end{figure}

The robot navigation tasks are borrowed from Safety Gym~\cite{ray2019benchmarking}, as shown in Fig. \ref{fig: safety gym snapshots}.
\textbf{PointGoal} and \textbf{CarGoal} require the agent to reach the goal while avoiding hazards, and \textbf{PointPush} and \textbf{CarPush} require the agent to push the box to the goal.
% There are 8 hazards with a radius of 0.2 and a goal with a radius of 0.3. The state includes the velocity of the robot, the position of the goal, and LiDAR point clouds of the hazards. The control inputs of the robots are the torques of their motors, controlling the motion of moving forward and turning for a Point robot, and the left and right wheels for a Car robot.
% There are 4 hazards with a radius of 0.1 and a goal with a radius of 0.3. The state further includes the position of the box.

\subsection{Results}
For grid world tasks, we adopt the framework of Q-learning~\cite{watkins1992q}, which uses a table to store the values of all state-action pairs.
Apart from FPI, which in this case we denote as \textbf{Q-learning-FPI}, we include another two algorithms for comparison: vanilla \textbf{Q-learning} and Q-learning with the Lagrange method, which we denote as \textbf{Q-learning-Lag}.
Both Q-learning-FPI and Q-learning-Lag maintain another Q table for storing cost values.
Q-learning-Lag selects the optimal action from a weighted sum of two Q tables, with the weight adjusted by dual ascent.
Q-learning-FPI uses the cost Q table to eliminate infeasible actions and selects the optimal one from the rest using the reward Q table.
% Vanilla Q-learning is included for inspecting the behavior of a constraint-agnostic policy.
Hyperparameters can be found in Appendix \ref{sec: hyperparameters}. 

For classic control tasks and robot navigation tasks, we compare FPI with (1) iterative unconstrained RL algorithms: SAC Lagrangian (\textbf{SAC-Lag})~\cite{ha2021learning} and SAC with CBF as constraint (\textbf{SAC-CBF}), and (2) constrained policy optimization methods: \textbf{CPO}~\cite{achiam2017constrained} and PPO Lagrangian (\textbf{PPO-Lag})~\cite{ray2019benchmarking}.
% For iterative unconstrained RL methods, we choose . 
The CBFs are handcrafted and described in Appendix \ref{sec: appx-cbf}.
% For constrained policy optimization methods, we choose .
% All hyperparameters can be found in Appendix \ref{sec: hyperparameters}. 

\subsubsection{Training curves}
\begin{figure*}[htbp]
    \centering
    \subfloat{\includegraphics[width=0.25\linewidth, trim=10 10 10 10, clip]{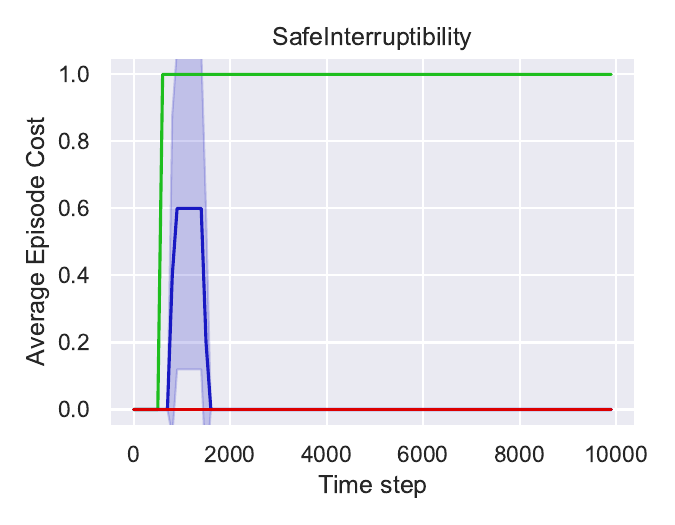}}
    \subfloat{\includegraphics[width=0.25\linewidth, trim=10 10 10 10, clip]{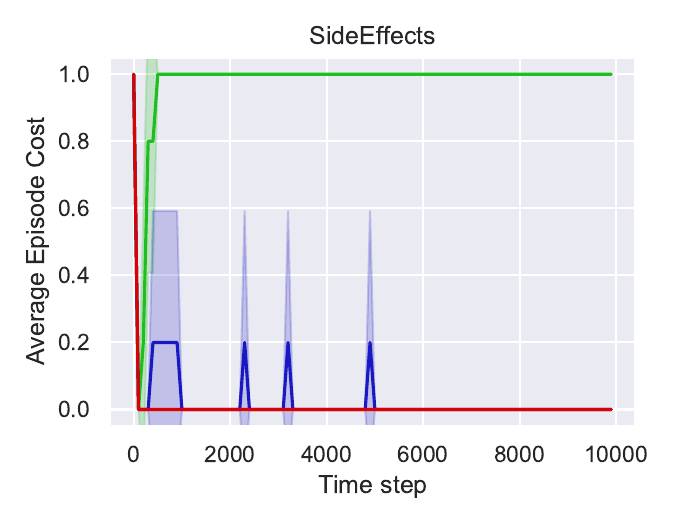}}
    \subfloat{\includegraphics[width=0.25\linewidth, trim=10 10 10 10, clip]{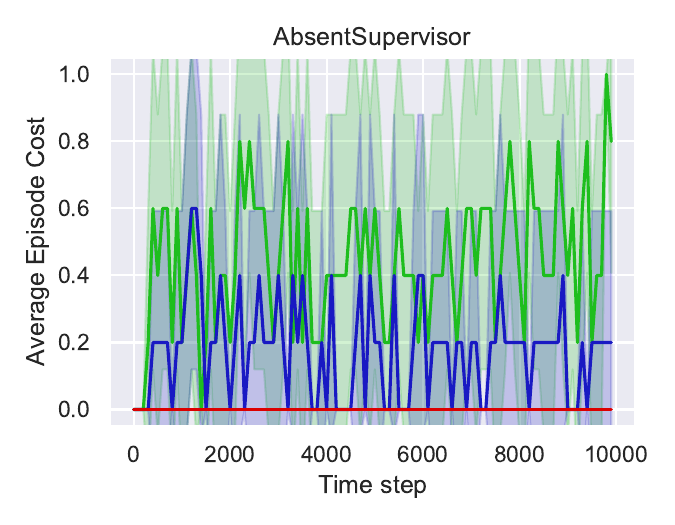}}
    \subfloat{\includegraphics[width=0.25\linewidth, trim=10 10 10 10, clip]{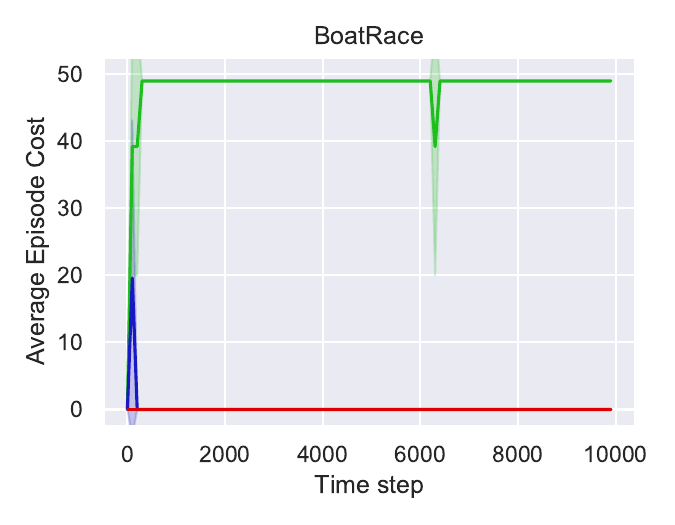}}\\
    \subfloat{\includegraphics[width=0.25\linewidth, trim=10 10 10 10, clip]{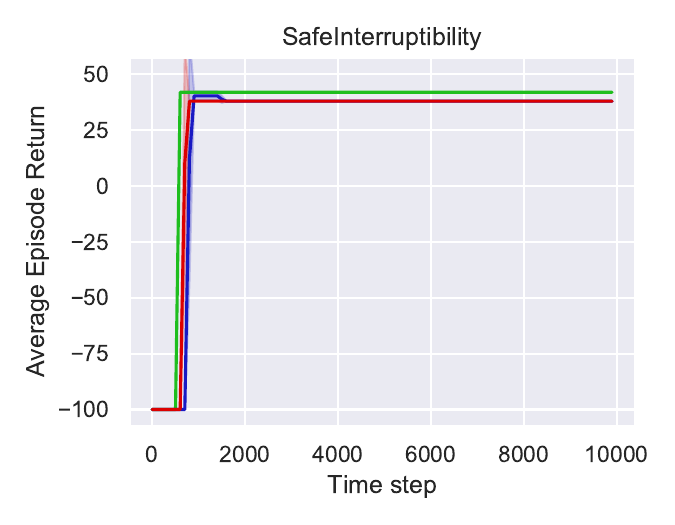}}
    \subfloat{\includegraphics[width=0.25\linewidth, trim=10 10 10 10, clip]{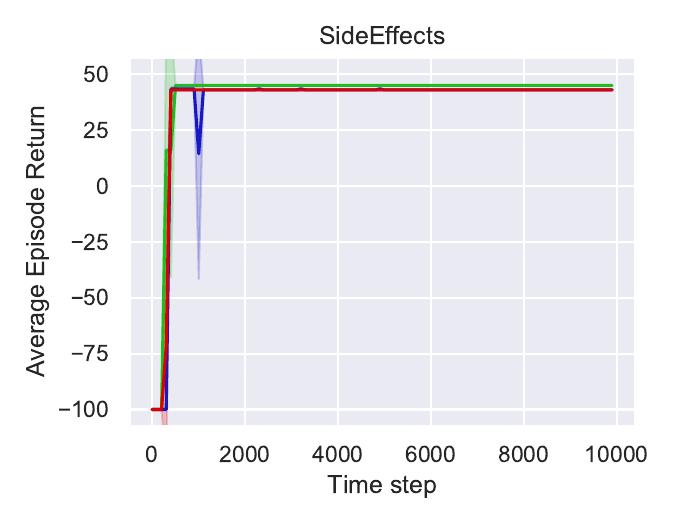}}
    \subfloat{\includegraphics[width=0.25\linewidth, trim=10 10 10 10, clip]{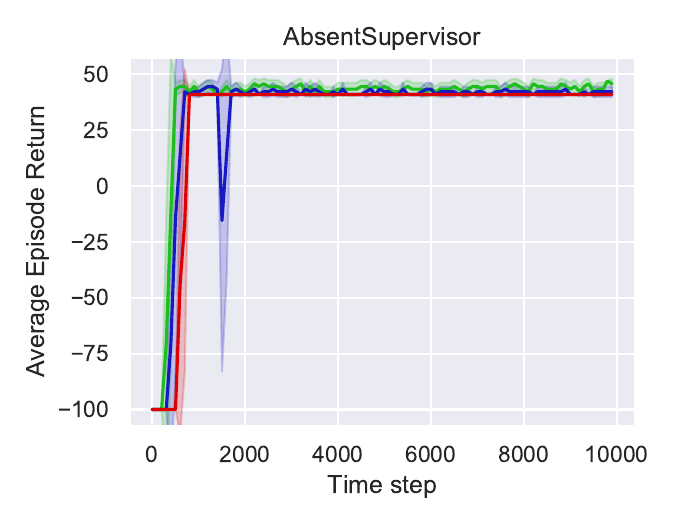}}
    \subfloat{\includegraphics[width=0.25\linewidth, trim=10 10 10 10, clip]{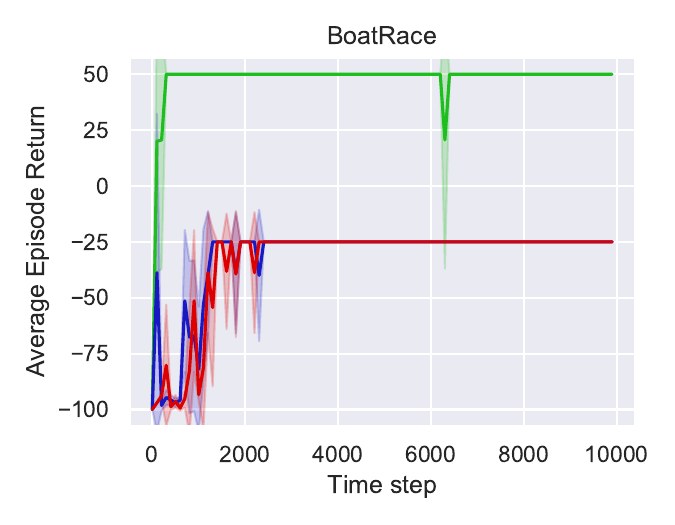}}\\
    \subfloat{\includegraphics[width=0.4\linewidth]{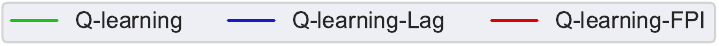}}
    \caption{Training curves on grid world tasks. The solid lines represent the mean and the shaded areas represent the 95\% confidence intervals over 5 seeds. The first row shows the average episode cost, and the second row shows the average episode return.}
    \label{fig: grid world training curves}
\end{figure*}

The training curves of average episode cost and average episode return on grid world tasks are shown in Fig. \ref{fig: grid world training curves}.
% Q-learning constantly violates the constraints, indicating that a reward-oriented policy is not safe on these tasks.
The superiority of FPI over the Lagrange method is evident: Q-learning-FPI quickly converges to zero constraint violation and optimal return across all tasks, while Q-learning-Lag exhibits slower convergence and significant oscillations.
The reason is that FPI immediately eliminates infeasible actions once they are identified by the feasibility function.
In contrast, the Lagrange method requires iteratively updating the multiplier until the cost Q value overweighs the reward Q value to eliminate infeasible actions.
% Although it is possible to adjust the initial value and learning rate of the multiplier to accelerate the convergence of the Lagrange method, such adjustments are highly task-specific and typically involve labor-intensive trial-and-error.

\begin{figure*}[htbp]
    \centering
    \subfloat{\includegraphics[width=0.25\linewidth, trim=10 10 10 10, clip]{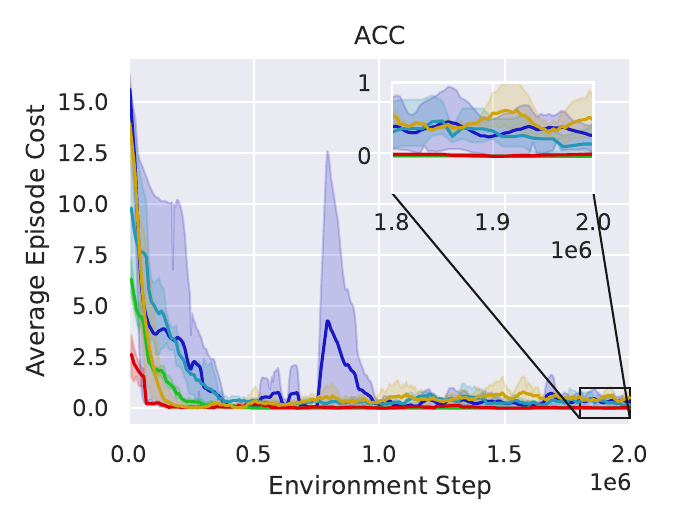}}
    \subfloat{\includegraphics[width=0.25\linewidth, trim=10 10 10 10, clip]{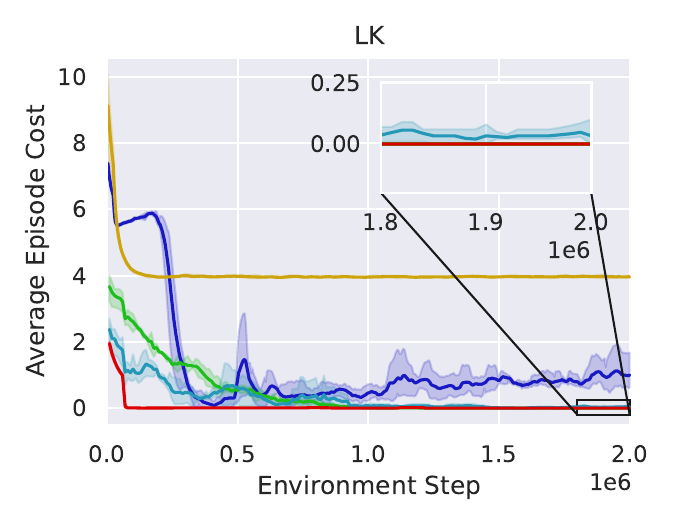}}
    \subfloat{\includegraphics[width=0.25\linewidth, trim=10 10 10 10, clip]{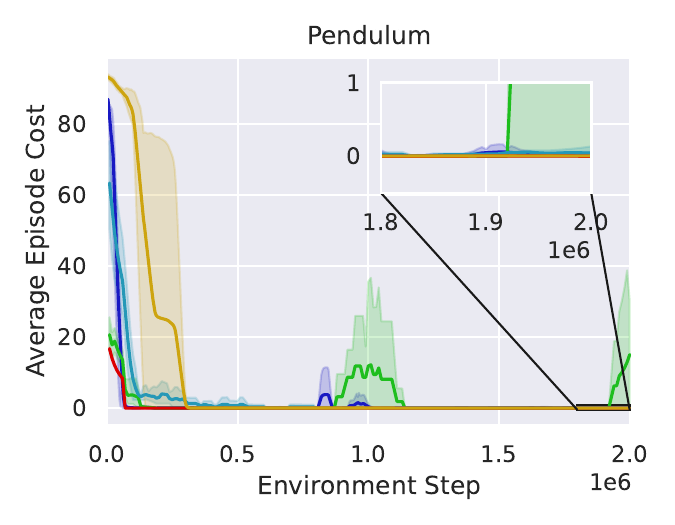}}
    \subfloat{\includegraphics[width=0.25\linewidth, trim=10 10 10 10, clip]{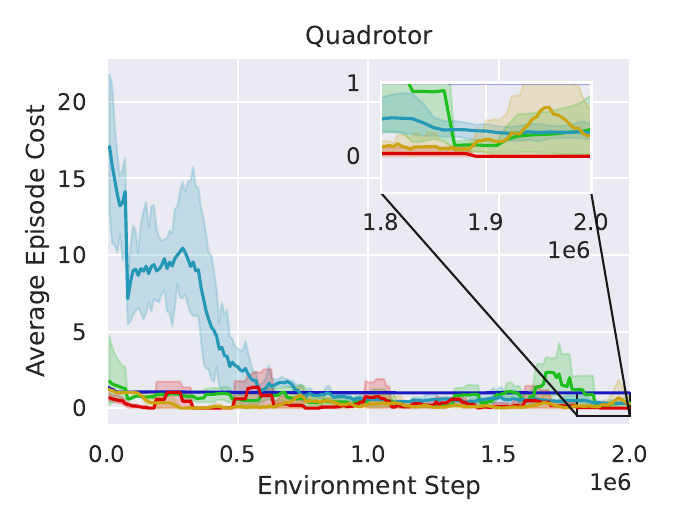}}\\
    \subfloat{\includegraphics[width=0.25\linewidth, trim=10 10 10 10, clip]{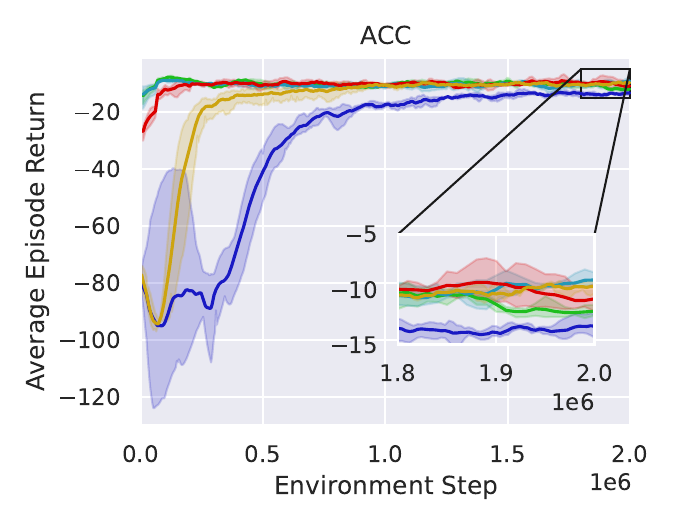}}
    \subfloat{\includegraphics[width=0.25\linewidth, trim=10 10 10 10, clip]{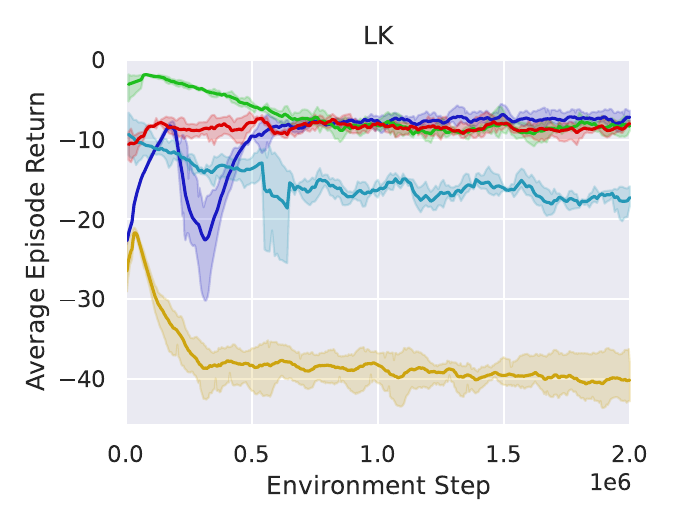}}
    \subfloat{\includegraphics[width=0.25\linewidth, trim=10 10 10 10, clip]{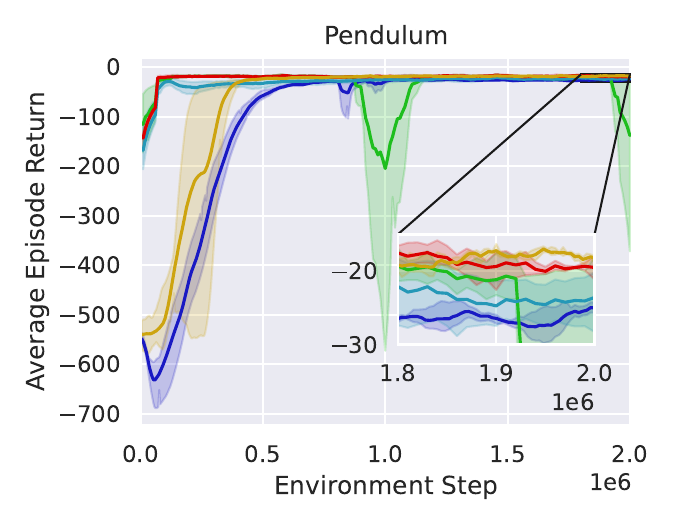}}
    \subfloat{\includegraphics[width=0.25\linewidth, trim=10 10 10 10, clip]{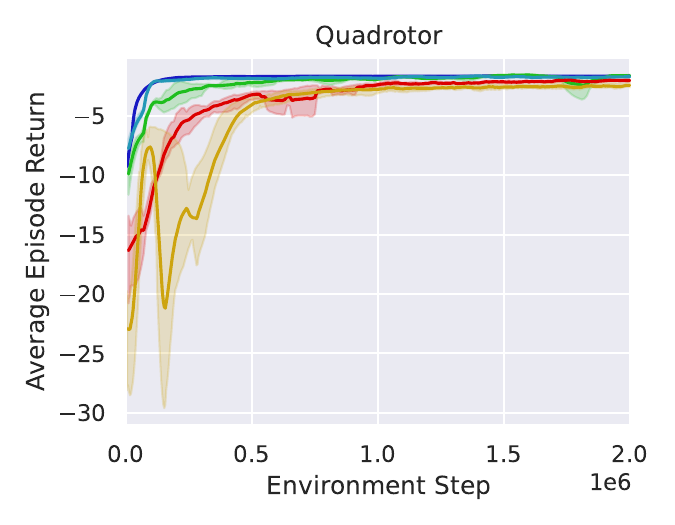}}\\
    \subfloat{\includegraphics[width=0.25\linewidth, trim=10 10 10 10, clip]{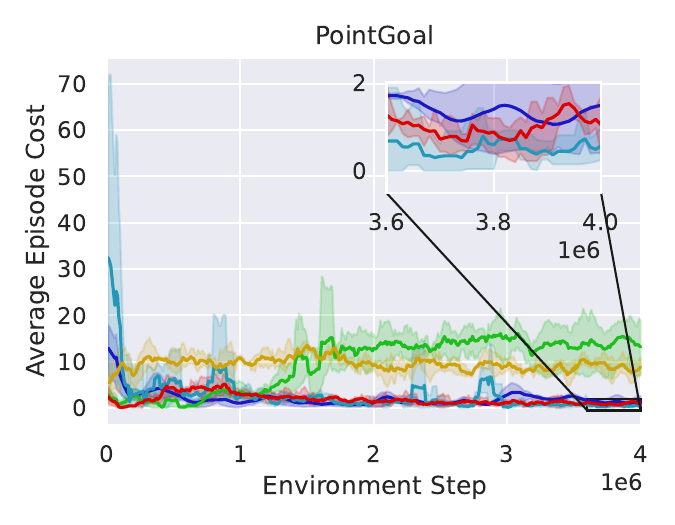}}
    \subfloat{\includegraphics[width=0.25\linewidth, trim=10 10 10 10, clip]{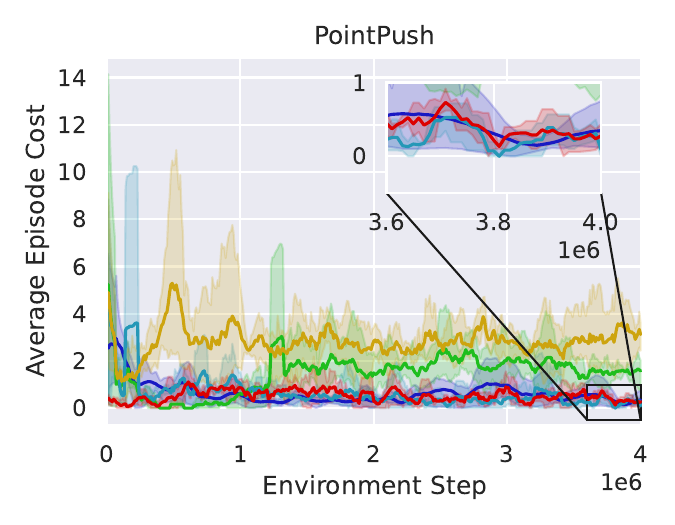}}
    \subfloat{\includegraphics[width=0.25\linewidth, trim=10 10 10 10, clip]{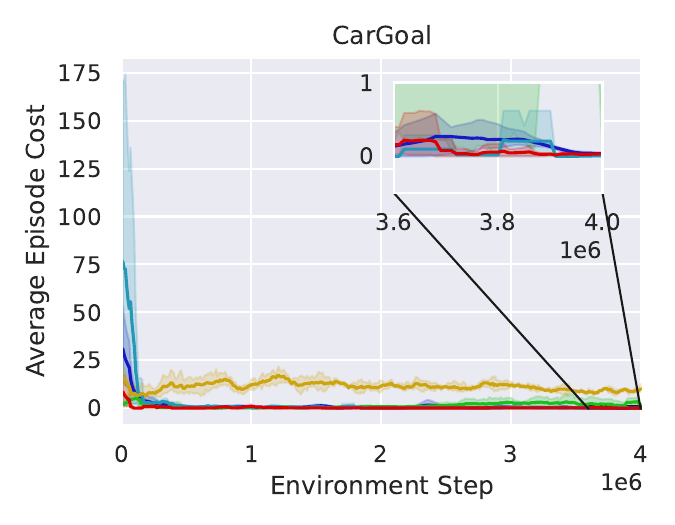}}
    \subfloat{\includegraphics[width=0.25\linewidth, trim=10 10 10 10, clip]{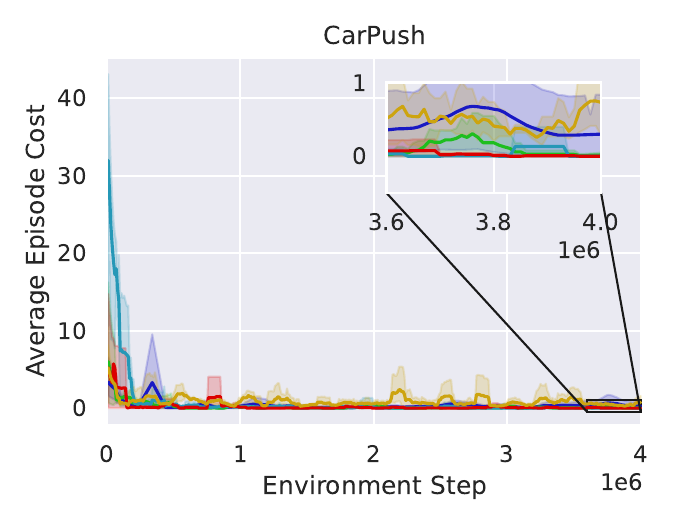}}\\
    \subfloat{\includegraphics[width=0.25\linewidth, trim=10 10 10 10, clip]{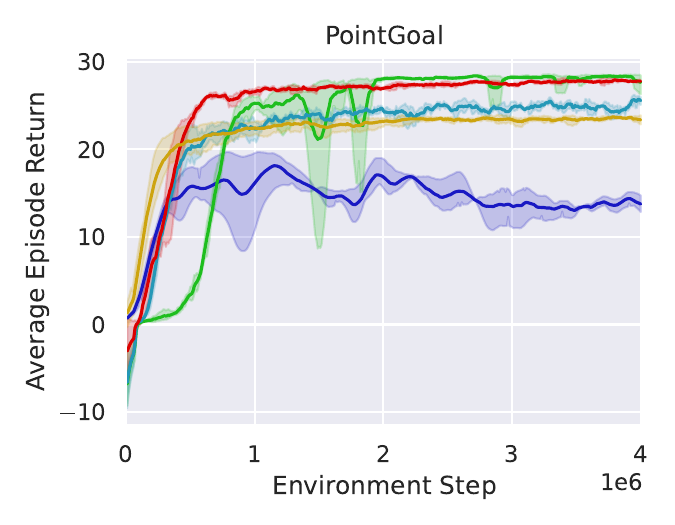}}
    \subfloat{\includegraphics[width=0.25\linewidth, trim=10 10 10 10, clip]{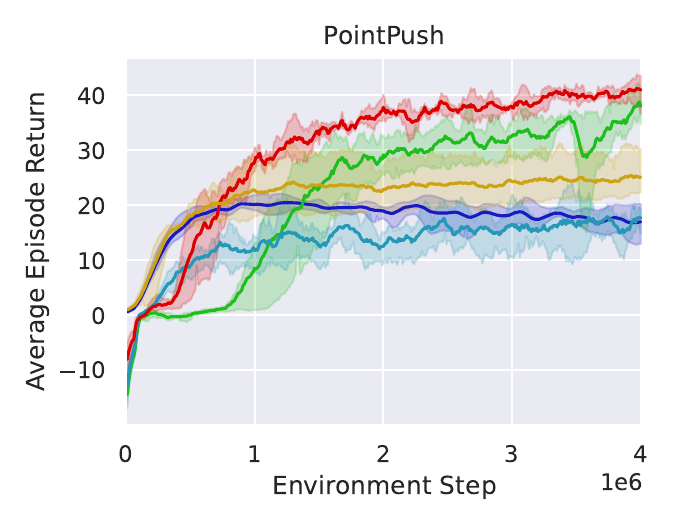}}
    \subfloat{\includegraphics[width=0.25\linewidth, trim=10 10 10 10, clip]{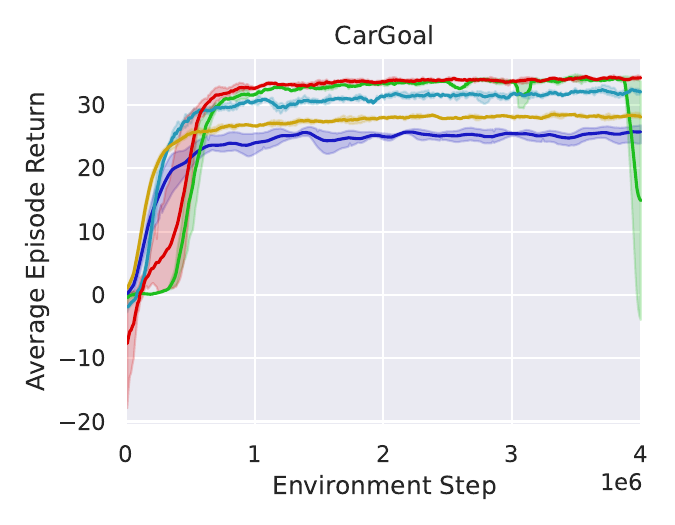}}
    \subfloat{\includegraphics[width=0.25\linewidth, trim=10 10 10 10, clip]{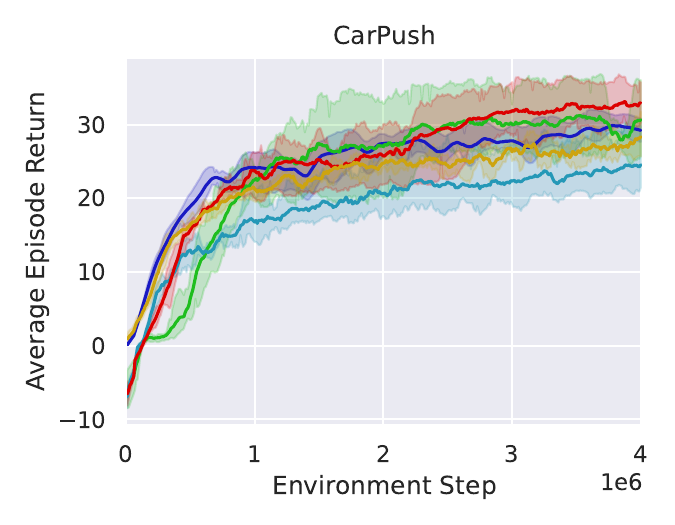}}\\
    \subfloat{\includegraphics[width=0.65\linewidth, trim=0 20 0 20, clip]{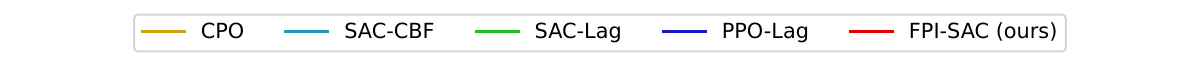}}
    \caption{Training curves on classic control and robot navigation tasks. The shaded areas represent the 95\% confidence intervals over 3 seeds.}
    \label{fig: class and robot training curves}
\end{figure*}

The training curves on classic control and robot navigation tasks are shown in Fig. \ref{fig: class and robot training curves}. 
FPI-SAC achieves comparable or higher performance than the baselines while keeping near-zero constraint violations on all tasks. %, demonstrating not only low but also stable episode cost curves.
SAC-CBF also performs fairly safe, but this comes at the cost of sacrificing optimality. %, which is the consequence of the overly restrictive constraint.
In comparison, the episode cost curves of other algorithms either fail to converge to zero or fluctuate severely. For example, CPO acts highly unsafely on LK and all robot navigation tasks except CarPush, so does SAC-Lag on two Point robot tasks, and PPO-Lag on LK and Quadrotor.
These are due to the impacts of CPO's infeasibility problem and the Lagrange multiplier method's lack of monotonicity guarantee. 
% Note that although PPO-Lag also performs well on robot navigation tasks in terms of safety, its returns on two Point robot tasks are low. Note also how the curves of SAC-Lag and PPO-Lag oscillate or even become worse during training, demonstrating the instability of the Lagrange multiplier method.
 
\subsubsection{Comparison with optimal policy}
For four classic control tasks, we use model predictive control (MPC) with a long enough horizon to approximate an optimal controller and compare the five algorithms with the MPC controller.
Evaluation metrics include constraint violation ratio $R_\text{vio}=N_\text{vio}/{N}$, and average normalized return $R_\text{norm}=(R_\text{alg}-R_\text{base})/(R_\text{MPC}-R_\text{base})-1$, where $N_\text{vio}$ is the number of episodes where the constraint is violated, and $N$ is the total number of episodes. We uniformly discretize the state space to form a grid of feasible initial states, each point of which yields an episode. $R_\text{alg}$, $R_\text{base}$ and $R_\text{MPC}$ are the average return across $N$ episodes of a certain algorithm, a random policy and an MPC controller, respectively.
The results are averaged across three random seeds, as listed in Table \ref{tab: classic control tasks results}.
We further apply an MPC controller without considering any constraint to show that the constraints do take effect.
Results show that FPI-SAC achieves zero constraint violation on all four tasks, while all other algorithms violate constraints on one or more tasks. Meanwhile, FPI-SAC ranks high in terms of return on all tasks.
% , especially ACC and Quadrotor, where FPI-SAC approaches the optimal solutions (The high returns of CPO on ACC come with constraint violations). On LK and Pendulum, FPI-SAC also achieves comparable returns to the best-performing algorithms.

\begin{table*}[htbp]
    \centering
    \resizebox{\textwidth}{!}{
    \begin{threeparttable}
    \caption{Constraint violation and normalized return on four classic control tasks.}
    \label{tab: classic control tasks results}
    \begin{tabular}{cr
    @{\hspace{1pt}$\pm$\hspace{1pt}}lr
    @{\hspace{1pt}$\pm$\hspace{1pt}}lr
    @{\hspace{1pt}$\pm$\hspace{1pt}}lr
    @{\hspace{1pt}$\pm$\hspace{1pt}}lr
    @{\hspace{1pt}$\pm$\hspace{1pt}}lr
    @{\hspace{1pt}$\pm$\hspace{1pt}}lr
    @{\hspace{1pt}$\pm$\hspace{1pt}}lr
    @{\hspace{1pt}$\pm$\hspace{1pt}}lr
    }
        \toprule
        & \multicolumn{4}{c}{ACC}
        & \multicolumn{4}{c}{LK}
        & \multicolumn{4}{c}{Pendulum}
        & \multicolumn{4}{c}{Quadrotor}
        \\
        \midrule
        & \multicolumn{2}{c}{$R_\text{vio}$}
        & \multicolumn{2}{c}{$R_\text{norm}$}
        & \multicolumn{2}{c}{$R_\text{vio}$}
        & \multicolumn{2}{c}{$R_\text{norm}$}
        & \multicolumn{2}{c}{$R_\text{vio}$}
        & \multicolumn{2}{c}{$R_\text{norm}$}
        & \multicolumn{2}{c}{$R_\text{vio}$}
        & \multicolumn{2}{c}{$R_\text{norm}$}
        \\
        \midrule
        CPO & $4.17$&$0.00$ & $0.17$&$0.14$ & $100.00$&$0.00$ & $-17.56$&$1.59$ & $0.00$&$0.00$ & $-0.04$&$0.01$ & $0.00$&$0.00$ & $-0.06$&$0.01$ \\ 
        SAC-Lag & $0.00$&$0.00$ & $-0.20$&$0.17$ & $0.00$&$0.00$ & $-0.22$&$0.06$ & $0.00$&$0.00$ & $-0.59$&$0.81$ & $8.89$&$3.85$ & $-5.81$&$1.55$ \\ 
        PPO-Lag & $2.78$&$2.41$ & $-1.54$&$0.44$ & $8.70$&$0.00$ & $-0.36$&$0.60$ & $2.90$&$2.51$ & $-0.62$&$0.22$ & $100.00$&$0.00$ & $-94.34$&$1.37$ \\
        SAC-CBF & $4.17$&$0.00$ & $-0.07$&$0.05$ & $0.00$&$0.00$ & $-3.84$&$0.34$ & $8.70$&$0.00$ & $-1.34$&$1.15$ & $31.11$&$6.29$ & $-28.43$&$5.37$ \\
        FPI-SAC (ours) & $0.00$&$0.00$ & $-0.01$&$0.01$ & $0.00$&$0.00$ & $-0.25$&$0.17$ & $0.00$&$0.00$ & $-0.10$&$0.02$ & $0.00$&$0.00$ & $-0.03$&$0.01$ \\ 
        MPC w/o cstr. & $20.83$&$0.00$ & $0.49$&$0.00$ & $43.48$&$0.00$ & $0.17$&$0.00$ & $17.39$&$0.00$ & $1.14$&$0.00$ & $100.00$&$0.00$ & $0.03$&$0.00$ \\
        \bottomrule
    \end{tabular}
    \begin{tablenotes}
    \item[1] The values denote mean$\pm$standard deviation.
    \item[2] MPC w/o cstr. refers to an MPC controller only maximizing return without considering constraints.
    \end{tablenotes}
    \end{threeparttable}}
\end{table*}

\subsubsection{Visualization}

\begin{figure*}[htbp]
    \centering
    \subfloat[ACC]{\label{fig:feasible region-ACC}\includegraphics[width=0.5\linewidth, trim=5 12 10 9, clip]{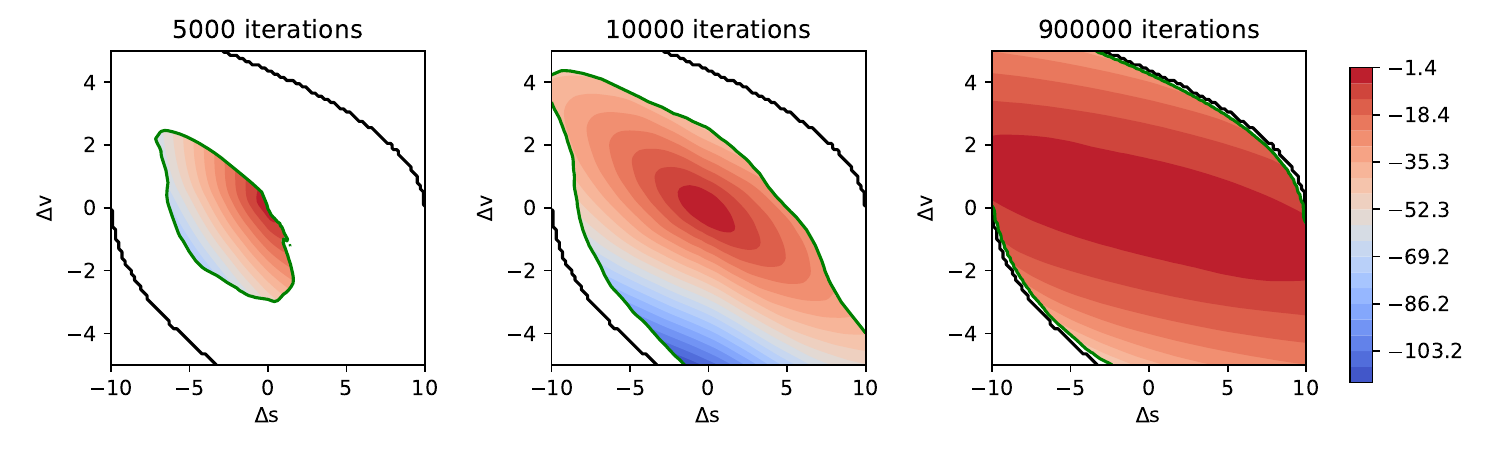}}
    \subfloat[LK]{\label{fig:feasible region-LK}\includegraphics[width=0.5\linewidth, trim=10 10 10 9, clip]{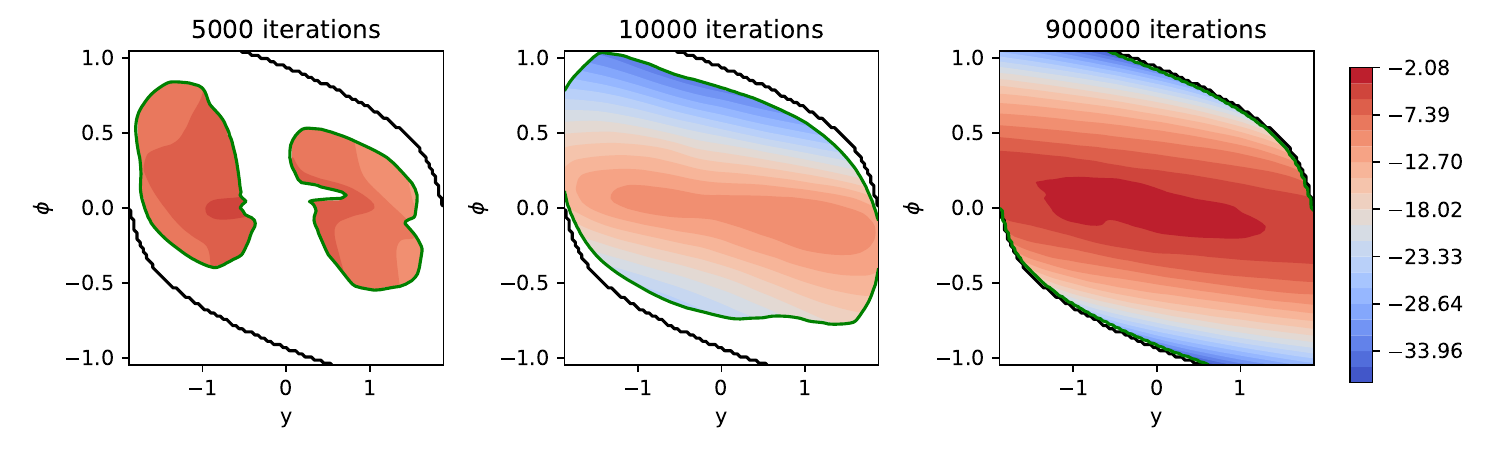}}
    \\
    \subfloat[Pendulum]{\label{fig:feasible region-pendulum}\includegraphics[width=0.5\linewidth, trim=10 12 10 9, clip]{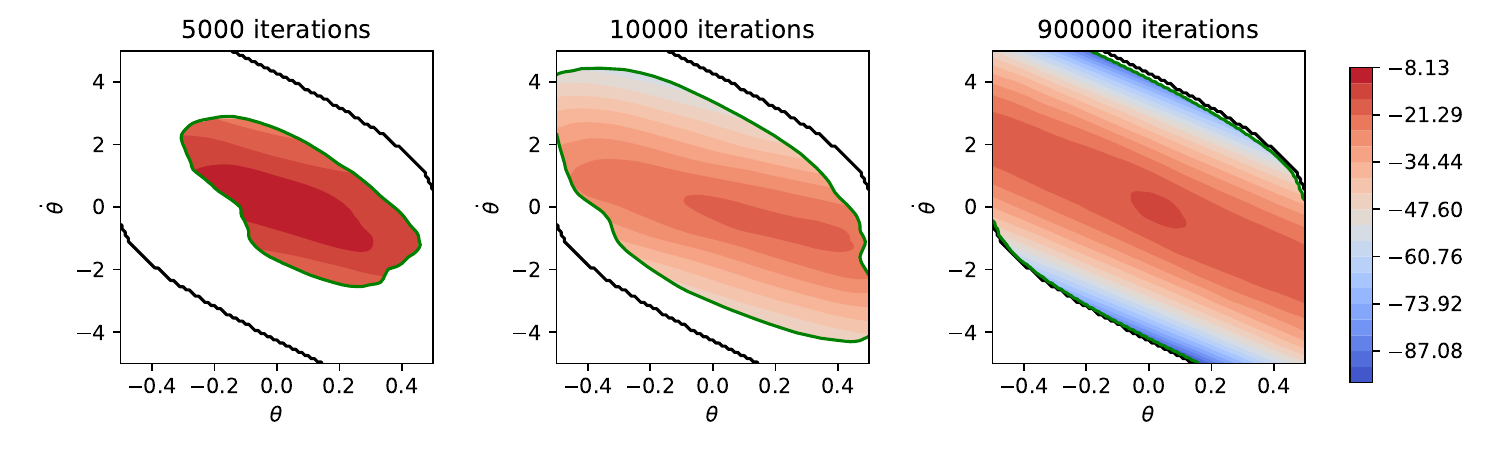}}
    \subfloat[Quadrotor]{\label{fig:feasible region-quadrotor}\includegraphics[width=0.5\linewidth, trim=10 12 10 9, clip]{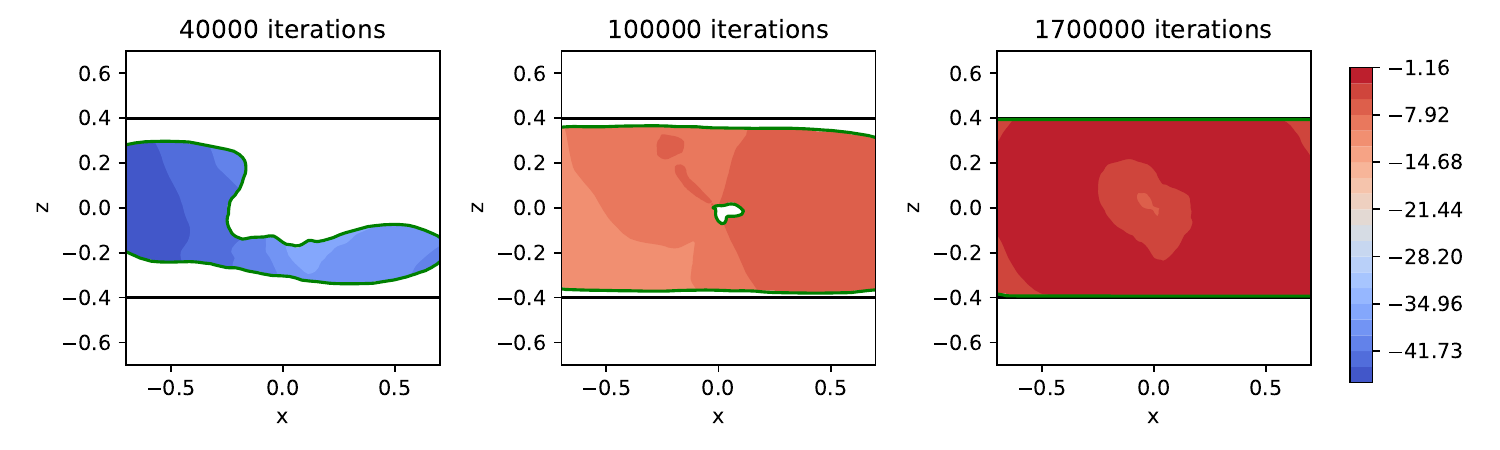}}
    \caption{Heat maps and zero-sublevel sets of the feasibility functions on four classic control tasks. The heat maps correspond to the state values. The black lines are boundaries of the maximum feasible regions, while the green ones around the colored regions are boundaries of zero contours of the feasibility functions.}
    \label{fig:feasible region}
\end{figure*}

To show the monotonic expansion of the feasible region and monotonic improvement of the state-value function, we visualize some intermediate training results of FPI-SAC, as shown in Fig. \ref{fig:feasible region}. The state values are presented as heat maps. The maximum feasible regions and the current feasible regions are outlined with black lines and green lines, respectively.
% We obtained the maximum feasible regions by applying the most conservative, safety-only policy to see if that could reverse the initial constraint-violating tendency. Take ACC as an instance. We use a policy that always takes the maximum deceleration when the initial $\Delta v < 0$ and a policy that always takes the maximum acceleration when the initial $\Delta v > 0$. If the constraint is still violated at the original side, that means the initial state is infeasible, otherwise, it is feasible.
In all tasks, FPI-SAC demonstrates monotonicity in terms of both the feasible region and state value, and the feasible regions gradually converge to the maximum ones. Note that the initial state values in Fig. \subref*{fig:feasible region-LK} and \subref*{fig:feasible region-pendulum} are high only because the value networks are initialized to produce near-zero outputs and haven't acquired to give the true values at this early stage.

\section{Conclusion}
This paper proposes FPI, the first foundational dynamic programming algorithm for safe RL.
FPI follows the ACS iteration scheme and iteratively performs PEV, RID, and region-wise PIM.
We prove that during the process of FPI, the feasible region monotonically expands, the state-value function monotonically increases within the feasible region, and both of them geometrically converge to their optima.
Based on FPI, we develop an algorithm with neural network approximation to address continuous state and action spaces.
Experiments show that our algorithm learns strictly safe and near-optimal policies with accurate feasible regions on low-dimensional tasks, and achieves lower constraint violations and better performance than existing algorithms on high-dimensional tasks.

\appendices

\section{Details of classic control tasks}
\label{sec: appx-env}
% In order to avoid confusion in symbols, in this section, we mark the states and actions in bold to emphasize that they are vectors.

In ACC, the system dynamics is
% a following vehicle tries to converge to and maintain a fixed distance with respect to a leading vehicle moving in a uniform motion.
% Both following and leading vehicles are modeled as point masses moving in a straight line. 
$$
\begin{pmatrix}
    \dot{x}_1  \\
    \dot{x}_2
\end{pmatrix}
=
\begin{pmatrix}
    x_2  \\
    0
\end{pmatrix}
+ 
\begin{pmatrix}
    0  \\
    -1
\end{pmatrix} \pmb u,
$$
where $\pmb x=\left[x_1,x_2\right]^\top \triangleq \left[\Delta s,\Delta v\right]^\top$, with $\Delta s=s-s_0$ standing for the difference between actual distance $s$ and expected distance $s_0$ between the two vehicles and $\Delta v$ standing for the relative velocity. The action $\pmb u\triangleq [a]$ is the acceleration of the following vehicle. 
The reward function is defined as $r\left(\pmb x,\pmb u\right) = -0.001{\Delta s}^2 - 0.01{\Delta v}^2 - a^2$, and the constraint function is $h(\pmb x) = |\Delta s| - \Delta s_\text{max}$.
% , which, since a large acceleration is penalized in the reward, will be violated by a performance-only policy. 

In LK, the state $\pmb x\triangleq \left[y, \varphi, v, \omega\right]^\top$ includes $y$ the vertical position, $\varphi$ the yaw angle, $v$ the lateral velocity and $\omega$ the yaw rate. The vehicle follows a 2DoF bicycle model with the action be its steering angle, i.e. $\pmb u\triangleq[\delta]$. 
The reward function penalizes stabilization errors and aggressive behaviour: $r\left(\pmb x,\pmb u\right) = -0.01y^2 - 0.01\varphi^2 - v^2 - \omega^2 - \delta^2$, and the constraint function is $h(\pmb x) = |y| - L/2$, where $L$ is the width of the lane.

Pendulum is borrowed from the RL benchmark Gymnasium~\cite{brockman2016openai} 
% with some slight changes. The state $\pmb x\triangleq[\theta, \dot{\theta}]^\top$ consists of the pendulum's angle w.r.t. vertical $\theta$ and angular velocity $\dot{\theta}$ and the action is $\pmb u\triangleq [\tau]$, representing the torque applied on the fixed end of the pendulum. Different from the Gymnasium version, the reward function is changed to $r\left(\pmb x,\pmb u\right) = -0.1\theta^2 - 0.01\dot{\theta}^2 - \tau^2$, and 
with an additional constraint $h(\pmb x) = |\theta| - \theta_\text{max} \le 0$.

In Quadrotor, the state is $\pmb x\triangleq [x, \dot{x}, z, \dot{z}, \theta, \dot{\theta}]^\top$, where $(x,z)$ is the position of the quadrotor on $xz$-plane, and $\theta$ is the pitch angle. The action of the system is $\pmb u\triangleq \left[T_1, T_2\right]^\top$, including the thrusts generated by two pairs of motors.
The reward is:
$$
\begin{aligned}
r\left(\pmb x,\pmb u\right) = 
&-\left\|\left[x-x_\text{ref}, z-z_\text{ref}\right]\right\|_2^2 - 0.1{\theta}^2- \dot{\theta}^2 \\
&- 0.1(T_1 - T_0)^2 - 0.1(T_2 - T_0)^2,
\end{aligned}
$$
where $(x_\text{ref},z_\text{ref})$ is the reference position the quadrotor is supposed to be at, and $T_0$ is the thrust needed for balancing the gravity. The reference position moves along a circle $x^2+z^2=0.5^2$ with a constant angular velocity.
The constraint function $h(\pmb x) = \max\{|z|-z_\text{max}, |\theta|-\theta_\text{max}, \text{err}-\text{err}_\text{max}\}$, where $\text{err} = \left\|\left[x-x_\text{ref}, z-z_\text{ref}\right]\right\|_2$, restricts the quadrotor to stay in a rectangular area.

\section{Handcrafted CBFs}
\label{sec: appx-cbf}
With a CBF $B(\pmb x): \mathcal{X}\to\mathbb{R}$, the constraint is constructed as $B(f(\pmb x,\pmb u))-B(\pmb x)\le -\lambda B(\pmb x)$, where $\lambda>0$ is a constant.

The CBF for ACC is
$$
B(\pmb x) = \left\{
    \begin{aligned}
        -10 + \Delta s + 4.5 \Delta v ~~~ \Delta v \ge 0\\
        -10 - \Delta s - 3.2 \Delta v ~~~ \Delta v < 0
    \end{aligned}
\right..
$$

The CBF for LK is
$$
B(\pmb x) = \left\{
    \begin{aligned}
        -L/2 + 1.8y + 3.2\varphi^2 + 0.5v + 0.8\omega ~~~ \phi \ge 0\\
        -L/2 - 1.8y - 3.2\varphi^2 - 0.5v - 0.8\omega ~~~ \phi < 0
    \end{aligned}
\right..
$$

The CBF for Pendulum is
$$
B(\pmb x) = \left\{
    \begin{aligned}
        -\theta_{\text{max}} + \theta + 0.3\dot{\theta} ~~~ \dot{\theta} \ge 0\\
        -\theta_{\text{max}} - \theta - 0.3\dot{\theta}~~~ \dot{\theta} < 0
    \end{aligned}
\right..
$$

The CBF for the Quadrotor is
$$
\begin{aligned}
    B(\pmb x) = -z_{\text{max}} +& 2.35z^2 \\
    +&\left\{
        \begin{aligned}   
            &2 \text{err}^2 + 0.2 \dot{z}    &0\le z \le z_{\text{max}}\\
            &2 \text{err}^2 - 0.2 \dot{z}    &-z_{\text{max}} \le z < 0 \\
            &2 \text{err}^2   &\text{others}
        \end{aligned}
    \right..
\end{aligned}
$$

The CBF for Safety Gym environments is $B(\pmb x) = r - d - 0.3\dot{d}$, where $r$ is the radius of hazards and $d$ is the distance to the center of the nearest hazard.

\section{Hyperparameters}
\label{sec: hyperparameters}

\begin{table}[htbp]
    \centering
    \caption{Hyperparameters for grid world tasks}
    \begin{tabular}{lc}
        \toprule
        Hyperparameter & Value \\
        \midrule
        Discount factor & 0.9 \\
        Learning rate & 0.5 \\
        Random exploration probability ($\epsilon$-greedy) & 0.1 \\
        % Multiplier initial value & 1.0 \\
        % Multiplier learning rate & 0.5 \\
        % Multiplier update delay & 10 \\
        \bottomrule
    \end{tabular}
\end{table}

\begin{table}[htbp]
    \centering
    \caption{Hyperparameters for continuous space tasks}
    \begin{tabular}{lc|c}
        \toprule
        Hyperparameter & \multicolumn{2}{c}{Value} \\
        & Classic & Safety Gym \\

        \midrule
        % \textit{Shared} \\
        Discount factor & \multicolumn{2}{c}{0.99} \\
        Number of hidden layers & \multicolumn{2}{c}{2} \\
        Number of hidden neurons & \multicolumn{2}{c}{256} \\
        Optimizer &  \multicolumn{2}{c}{Adam($\beta_1$=0.99,$\beta_2$=0.999)}\\

        % \midrule
        % \textit{SAC-related} \\
        Activation function & \multicolumn{2}{c}{ReLU} \\
        Target entropy & \multicolumn{2}{c}{-dim($\mathcal{U}$)} \\
        Initial temperature & \multicolumn{2}{c}{1.0} \\
        Target smoothing coefficient & \multicolumn{2}{c}{0.005} \\
        Learning rate & \multicolumn{2}{c}{1e-4} \\
        Batch size & 256 & 1024 \\
        Replay buffer size & $2\times10^6$ & $4\times10^6$ \\
        
        % \midrule
        % \textit{Lagrange-related} \\
        % \quad Multiplier initial value & \multicolumn{2}{c}{1.0} \\
        % \quad Multiplier learning rate & \multicolumn{2}{c}{1e-4} \\
        % \quad Multiplier update delay & \multicolumn{2}{c}{10} \\

        % \midrule
        % \textit{CBF-related} \\
        % \quad $\lambda$ & \multicolumn{2}{c}{0.1} \\

        % \midrule
        % \textit{FPI-related (ours)} \\
        Feasibility threshold $(p)$ & \multicolumn{2}{c}{0.1} \\
        Initial $t$ & \multicolumn{2}{c}{1.0} \\
        $t$ increase factor & \multicolumn{2}{c}{1.1} \\
        $t$ update delay & \multicolumn{2}{c}{10000} \\
        \bottomrule
    \end{tabular}
    \label{tab: hyperparameters off-policy}
\end{table}

\section*{References}
\bibliographystyle{IEEEtran}
\bibliography{ref}

\end{document}